\PassOptionsToPackage{table}{xcolor}
\documentclass{bmvc2k}

\title{Robust and Label-Efficient Deep Waste Detection}

\addauthor{Hassan Abid}{hassan.abid@mbzuai.ac.ae}{\ensuremath{\dagger},1}
\addauthor{Khan Muhammad}{khan.muhammad@ieee.org}{\ensuremath{\dagger},2}
\addauthor{Muhammad Haris Khan}{muhammad.haris@mbzuai.ac.ae}{1}

\addinstitution{
Mohamed Bin Zayed University of Artificial Intelligence,\\
Abu Dhabi, UAE
}
\addinstitution{
 Sungkyunkwan University,\\
  Seoul, South Korea
}

\runninghead{ABID \textit{et al.}}{Robust and Label-Efficient Deep Waste Detection}

\def\etal{\emph{et al}\bmvaOneDot}

\usepackage{graphicx}
\usepackage{multirow}
\usepackage{booktabs}
\usepackage{float}
\usepackage{colortbl}
\usepackage{ragged2e}
\usepackage{algorithm}
\usepackage{algpseudocode}
\usepackage{tabularx,array,booktabs,xurl}
\definecolor{darkgreen}{rgb}{0,0.5,0}
\definecolor{mygray}{rgb}{.95,.95,.95}
\begin{document}

\maketitle

\begin{abstract}
Effective waste sorting is critical for sustainable recycling, yet AI research in this domain continues to lag behind commercial systems due to limited datasets and reliance on legacy object detectors. In this work, we advance AI-driven waste detection by establishing strong baselines and introducing an ensemble-based semi-supervised learning framework. We first benchmark state-of-the-art Open-Vocabulary Object Detection (OVOD) models on the real-world ZeroWaste dataset, demonstrating that while class-only prompts perform poorly, LLM-optimized prompts significantly enhance zero-shot accuracy. Next, to address domain-specific limitations, we fine-tune modern transformer-based detectors, achieving a new baseline of 51.6 mAP. We then propose a soft pseudo-labeling strategy that fuses ensemble predictions using spatial and consensus-aware weighting, enabling robust semi-supervised training. Applied to the unlabeled ZeroWaste-s subset, our pseudo-annotations achieve performance gains that surpass fully supervised training, underscoring the effectiveness of scalable annotation pipelines. Our work contributes to the research community by establishing rigorous baselines, introducing a robust ensemble-based pseudo-labeling pipeline, generating high-quality annotations for the unlabeled ZeroWaste-s subset, and systematically evaluating OVOD models under real-world waste sorting conditions. Our code is available at: \href{https://github.com/h-abid97/robust-waste-detection}{GitHub Repository}.
\end{abstract}

\section{Introduction}
\label{sec:intro}
The global waste crisis, driven by urbanization and increased consumption, poses a critical environmental, health, and economic challenge. The generation of municipal solid waste is projected to grow from 2.01 billion tonnes in 2016 to 3.40 billion tonnes by 2050~\cite{kaza2018waste}. Alarmingly, only 13.5\% of waste is recycled, while over 33\% is improperly managed, causing severe global challenges \cite{triassi2015environmental, kaza2018waste}. Improving recycling efficiency is essential for the advancement of several Sustainable Development Goals (SDGs) of the United Nations~\cite{kopecka2024role, un_sdg12}. Developing intelligent and scalable waste sorting technologies has therefore become a key priority for global sustainability efforts.

Material Recovery Facilities (MRFs) play a central role by sorting waste into recyclable streams such as plastics, metals, paper, and cardboard~\cite{britannica_mrf, bashkirova2022zerowaste}. However, despite utilizing machinery alongside manual labor~\cite{gundupalli2017review}, recycling rates remain low, with less than 35\% of recyclable waste recovered in the United States as of 2018~\cite{epa2018}. Conventional MRF operations suffer from inefficiencies and high material losses, discarding up to 20\% of recyclables~\cite{amp2022robotics}, and expose workers to hazardous conditions including sharp objects, toxic substances, and medical waste~\cite{gundupalli2017review}. Moreover, the cluttered, overlapping, and deformable nature of waste streams makes automated detection and sorting particularly challenging. To address these limitations, recent advances have integrated Artificial Intelligence (AI) and robotics into MRF workflows, with object detection models emerging as a practical foundation for scalable, real-time sorting under complex conditions~\cite{zhou2023construction, seredkin2019automated}.
Although several commercial companies~\cite{amp_robotics, zen_robotics, waste_robotics} have developed effective AI-powered waste sorting solutions, their models and datasets remain proprietary, limiting broader research progress. In contrast, most academic studies rely on outdated object detection baselines and simplistic datasets collected in controlled environments~\cite{yang2016classification, haamer2020wadeai, proenca2020taco, serezhkin2020drinkingwaste, datacluster2021domestictrash}, which fail to reflect the real-world challenges of industrial waste streams. Consequently, research in AI-driven waste detection significantly lags behind commercial advances, hindering reproducibility and scalability. Additionally, recent progress in OVOD~\cite{minderer2022simple, minderer2023scaling, liu2024grounding, Cheng_2024_CVPR}, which leverages vision-language models (VLMs) to generalize beyond fixed category sets, offers a promising avenue for improving adaptability. However, the effectiveness of OVOD models in complex industrial waste sorting scenarios remains largely unexplored. 

In this work, we make four key contributions: \textbf{(1)} We develop an ensemble-based pseudo-labeling pipeline to generate high-quality annotations for the previously unlabeled ZeroWaste-s dataset, enabling scalable benchmarking and reducing reliance on costly manual labeling. \textbf{(2)} We conduct a comprehensive zero-shot evaluation of state-of-the-art OVOD models on real-world waste sorting data, highlighting their strengths and limitations in cluttered, deformable environments. \textbf{(3)} We design a robust semi-supervised learning framework that leverages ensemble-based pseudo-labels to substantially improve detection performance through large-scale training on unlabeled data. \textbf{(4)} We establish new strong baselines for waste detection by fine-tuning advanced object detectors, providing critical benchmarks to guide future research in AI-driven waste recovery.

\vspace{-0.5em}
\section{Related Work}
\label{sec:related}
\textbf{Deep Learning for Waste Recognition.} Deep learning has been widely adopted for waste recognition tasks. Early works focused on image-level classification, utilizing convolutional neural networks (CNNs) trained on datasets such as TrashNet~\cite{yang2016classification} and TACO~\cite{proenca2020taco}. RecycleNet~\cite{bircanoglu2018recyclenet} and Vo \etal~\cite{vo2019trash} improved classification using DenseNet and ResNeXt architectures, while later studies explored lightweight models~\cite{feng2022intelligent, tian2023garbage}, metadata integration~\cite{sun2020thanosnet}, and attention mechanisms~\cite{liu2022depthwise, liu2022image}. However, these methods assume isolated objects in uncluttered settings, limiting their applicability to real-world MRFs characterized by severe clutter, deformation, and occlusion. To enable robotic sorting, research shifted to object detection frameworks capable of simultaneous classification and localization. Faster R-CNN~\cite{ren2016fasterrcnnrealtimeobject, mengistu2017smart} and YOLO variants~\cite{redmon2016lookonceunifiedrealtime, liu2018research, lin2021yolo} were adapted for waste detection, with lightweight improvements for real-time inference~\cite{ma2020lightweight, zhou2021towards, carolis2020yolo, xia2024yolo}. However, evaluations largely remained on controlled datasets. We address this limitation by benchmarking modern state-of-the-art detectors under realistic MRF conditions using the ZeroWaste dataset~\cite{bashkirova2022zerowaste}.

\noindent\textbf{Waste Recognition Datasets.} Public datasets for waste recognition vary widely in complexity and realism. Early datasets~\cite{yang2016classification, mittal2016gini, martin_recycling, proenca2020taco, sousa2019automation, koskinopoulou2021robotic} feature isolated objects, synthetic imagery, or simplified backgrounds that fail to capture the complexities of real-world settings. Recent datasets improve realism by collecting data in operational facilities~\cite{yudin2024hierarchical, 10801797}. However, they are not intended for the detection task.
ZeroWaste~\cite{bashkirova2022zerowaste} is collected from a full-scale MRF, offering the most comprehensive benchmark for industrial waste detection, with over 27,000 annotated instances across 4600 images and 6000 unlabeled frames. ZeroWaste stands out for its scale, strong emphasis on detection, and structured support for fully and semi-supervised learning. We adopt it as the most suitable dataset for advancing robust, scalable object detection in real-world waste sorting.

\noindent\textbf{Open-Vocabulary Object Detection.} OVOD~\cite{zareian2021open,zhou2022detectingtwentythousandclassesusing} enables recognition of object categories not seen during training by leveraging VLMs such as CLIP \cite{radford2021learningtransferablevisual} and ALIGN~\cite{jia2021scalingvisualvisionlanguagerepresentation}.
Recent approaches improve region-text alignment for open-set detection: OWL-ViT~\cite{minderer2022simple, minderer2023scaling} uses vision transformers trained on aligned image-text pairs; Grounding DINO~\cite{liu2024grounding} introduces language-guided detection with grounded pretraining and a strong transformer-based architecture; and YOLO-World~\cite{Cheng_2024_CVPR} extends the YOLO family for real-time OVOD via a contrastive region-text loss and vision-language path aggregation. While these models perform well on general-purpose benchmarks like COCO~\cite{lin2014microsoft}, LVIS~\cite{gupta2019lvis}, and Objects365~\cite{shao2019objects365}, their effectiveness under the severe occlusion, deformation, and clutter typical of waste sorting environments remains largely untested. We address this gap by systematically benchmarking OVOD models under domain-specific industrial conditions.

\noindent\textbf{Ensemble-Based Pseudo-Labeling.} Pseudo-labeling (PL) reduces manual annotation costs in object detection~\cite{sohn2020simple, liu2021unbiasedteacher, xu2021softteacher}. Ensemble-based approaches~\cite{zhou2021instant, liu2022dense} further enhance pseudo-label quality by aggregating multiple predictors, mitigating errors from individual models. Most prior ensembling efforts~\cite{sohn2020simple, liu2021unbiasedteacher, xu2021softteacher, zhou2021instant, liu2022dense} focus on clean datasets e.g., COCO~\cite{lin2014microsoft} and VOC~\cite{Everingham10}, with little exploration under dense clutter, deformation, and occlusion. We develop an ensemble-based PL pipeline for MRF waste detection, improving pseudo-label quality on the ZeroWaste-s subset and enabling scalable semi-supervised training.

\noindent\textbf{Semi-Supervised Object Detection.} Semi-Supervised Object Detection (SSOD) boosts detection performance by exploiting unlabeled data. Methods such as STAC~\cite{sohn2020simple}, Unbiased Teacher~\cite{liu2021unbiasedteacher}, Soft Teacher~\cite{xu2021softteacher}, and Dense Teacher~\cite{liu2022dense} use teacher-student architectures and consistency regularization. Although SSOD has shown strong results on benchmarks like COCO and VOC, it remains underexplored in dense industrial waste scenarios. We leverage ensemble-generated pseudo-labels for semi-supervised fine-tuning, establishing stronger baselines for waste recovery tasks.

\vspace{-1em}
\section{Our Approach}
\label{sec:methodology}
This section begins with describing the ZeroWaste dataset~\cite{bashkirova2022zerowaste} and its challenges, followed by a comprehensive zero-shot evaluation of state-of-the-art open-vocabulary detectors. We then establish new supervised baselines through fine-tuning the latest detectors on ZeroWaste-f, and finally, we introduce a semi-supervised framework that leverages ensemble-based soft pseudo-labeling to exploit the unlabeled ZeroWaste-s subset.

\vspace{-0.5em}
\subsection{Dataset}
\begin{figure}[htb]
  \centering
  \setlength{\tabcolsep}{10pt}
  \begin{tabular}{cc}
    \includegraphics[width=0.45\textwidth]{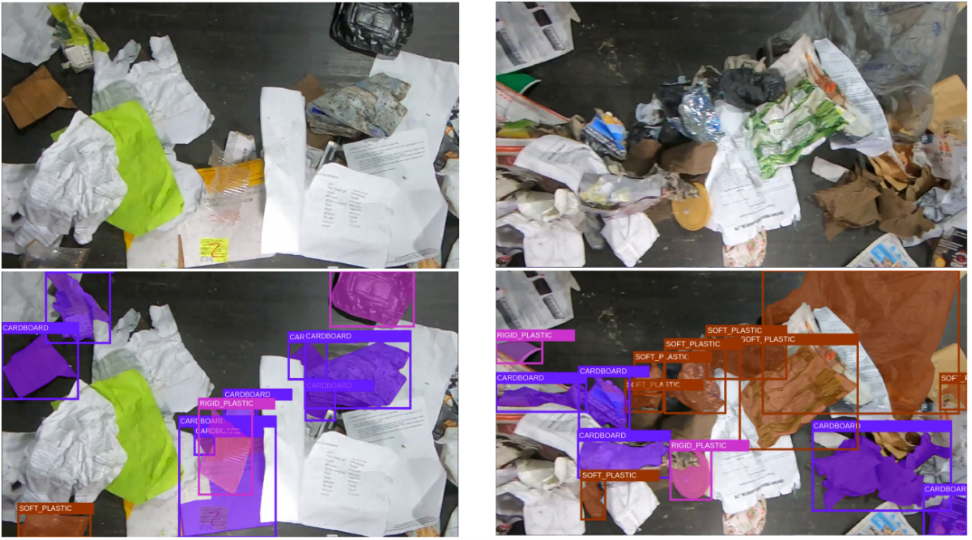} &
    \includegraphics[width=0.45\textwidth]{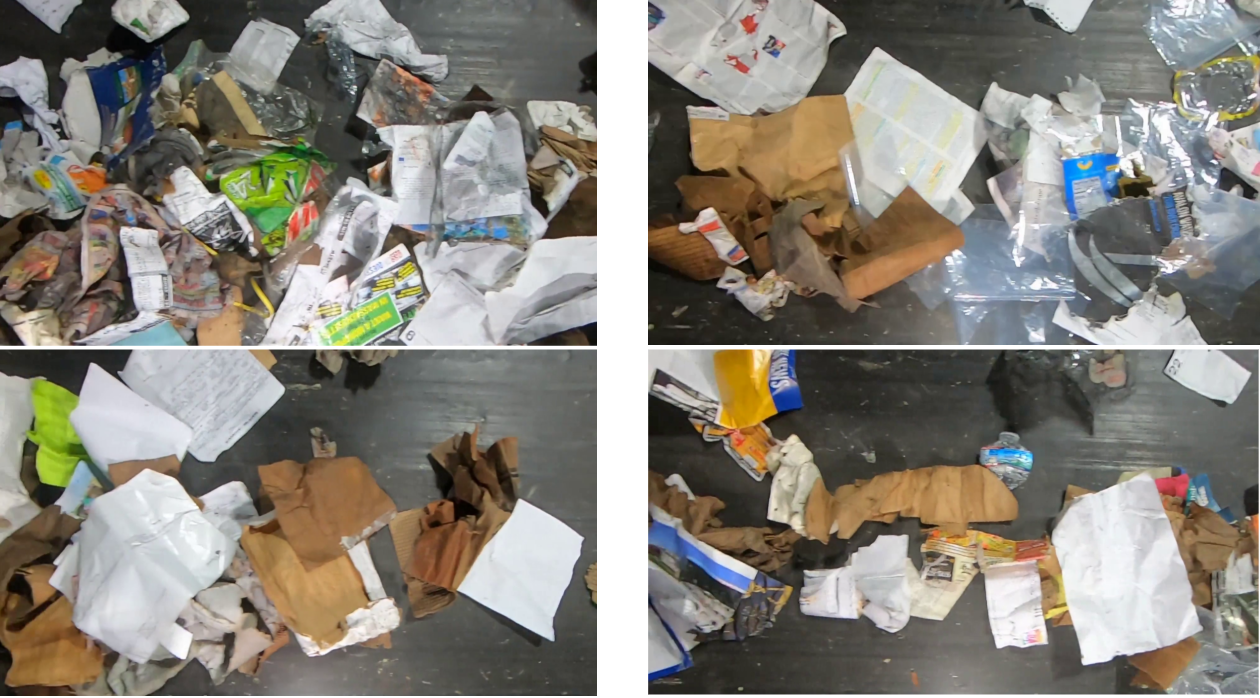} \\
    (a) ZeroWaste-f & (b) ZeroWaste-s
  \end{tabular}
  \vspace{-0.5em}
  \caption{Visual examples from the ZeroWaste dataset. (a) Sample images from ZeroWaste-f (top) with ground-truth annotations (bottom) for four recyclable categories: cardboard, soft plastic, rigid plastic, and metal. (b) Sample images from the unlabeled ZeroWaste-s subset, designed to support semi-supervised learning under the same real-world conditions.}
  \label{fig:two_subfigs}
\end{figure}

All experiments are conducted on the ZeroWaste dataset, collected in an operational MRF using high-resolution (1920×1080) overhead imagery that mimics real-world industrial sorting systems. Compared to conventional datasets, ZeroWaste introduces domain-specific challenges such as severe occlusions, deformations, cluttered backgrounds, and extreme class imbalance. The dataset comprises two subsets: \textbf{ZeroWaste-f}, which contains 4,503 labeled images with bounding box annotations for four recyclable categories—\textit{cardboard}, \textit{soft plastic}, \textit{rigid plastic}, and \textit{metal}; and \textbf{ZeroWaste-s}, which includes 6,212 unlabeled images captured under the same conditions to support semi-supervised learning. Fig.~\ref{fig:two_subfigs} shows examples from both subsets.
A particularly challenging aspect is the class imbalance: \textit{cardboard} accounts for over 66\% of annotations, whereas \textit{metal} comprises less than 2\%. With many frames containing 15+ objects, the dataset presents high clutter and detection difficulty. These characteristics make ZeroWaste a rigorous benchmark for evaluating object detectors in unconstrained environments. While the dataset defines a fixed set of categories, it enables evaluating open-vocabulary detectors in the zero-shot setting, assessing their ability to localize and recognize unseen categories without any category-specific training.

To contextualize model performance, the dataset authors also reported supervised baselines using CNN-based detectors. As shown in Table~\ref{tab:zerowaste_baselines}, TridentNet~\cite{li2019scale} achieved the best performance (mAP 24.2), outperforming RetinaNet~\cite{lin2017focal} and Mask R-CNN~\cite{he2017mask}. However, all models consistently scored poorly on detection metrics, highlighting the real-world difficulty of the dataset and motivating the need for updated baselines with modern architectures.

\begin{table}[htb]
\centering
\scriptsize
\renewcommand{\arraystretch}{1.0}

\begin{minipage}[t]{0.45\textwidth}
    \centering
    \setlength{\tabcolsep}{3pt}
    \renewcommand{\arraystretch}{1.3}
    \begin{tabular}{lcccccc}
        \toprule
        \textbf{Model} & \textbf{mAP} & \textbf{mAP50} & \textbf{mAP75} & \textbf{mAPs} & \textbf{mAPm} & \textbf{mAPl} \\
        \midrule
        RetinaNet   & 21.0 & 33.5 & 22.2 & 4.3 & 9.5 & 22.7 \\
        Mask R-CNN  & 22.8 & 34.9 & 24.4 & 4.6 & 10.6 & 25.8 \\
        TridentNet  & \textbf{24.2} & \textbf{36.3} & \textbf{26.6} & \textbf{4.8} & \textbf{10.7} & \textbf{26.1} \\
        \bottomrule
    \end{tabular}
\end{minipage}
\hspace{0.1\textwidth} 
\begin{minipage}[t]{0.42\textwidth}
    \vspace{-4.3em}
    \caption{Performance of CNN detectors (RetinaNet, Mask R-CNN, TridentNet) fine-tuned on ZeroWaste-f and evaluated on its test set, as reported by Bashkirova et al.~\cite{bashkirova2022zerowaste}.}
    \label{tab:zerowaste_baselines}
\end{minipage}
\end{table}

\vspace{-2.0em}
\subsection{Zero-Shot OVOD Baselines}
To establish baseline performance for OVOD in the ZeroWaste dataset, we evaluated three state-of-the-art models: Grounding DINO~\cite{liu2024grounding}, OWLv2~\cite{minderer2023scaling}, and YOLO-World~\cite{Cheng_2024_CVPR} in a zero-shot setting. These models were selected for their complementary architectures and proven capabilities. Grounding DINO achieves strong zero-shot grounding via cross-modality fusion and has set state-of-the-art benchmarks on COCO~\cite{lin2014microsoft} and LVIS~\cite{gupta2019lvis}; OWLv2 improves open-vocabulary performance on rare categories using large-scale self-training~\cite{minderer2023scaling}; and YOLO-World combines vision-language modeling with real-time efficiency, making it suitable for industrial deployment~\cite{Cheng_2024_CVPR}.

\begin{figure}[!htp]
  \centering
  \setlength{\tabcolsep}{4pt}
  \begin{tabular}{cc}
    \includegraphics[width=0.46\textwidth]{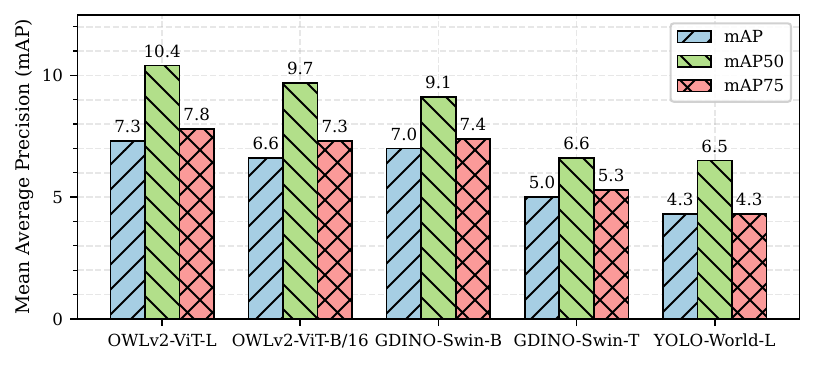} &
    \includegraphics[width=0.46\textwidth]{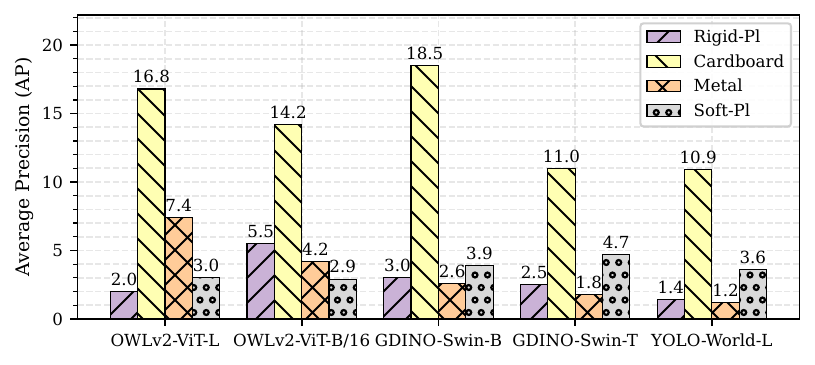} \\
    (a) Overall mAP Values & (b) Category-wise AP Values 
  \end{tabular}
  \vspace{0.6em}
    \caption{
    Zero-shot detection performance on the ZeroWaste-f test set using class-only prompts. 
    \textbf{(a)} Overall mAP scores remain low across models, with OWLv2-ViT-L achieving the best mAP (7.3). 
    \textbf{(b)} Category-wise AP reveals strong performance on cardboard, while soft plastic and metal suffer due to transparency and reflectivity, highlighting challenges.
    }
  \label{fig:ovod_classonly_results} \vspace{-1em}
\end{figure}

\vspace{+0.8em}
\noindent\textbf{Baseline Evaluation with Class-Only Prompts.} To assess raw zero-shot performance, we provide each model with simple category-level prompts, namely \textit{"cardboard"}, \textit{"soft plastic"}, \textit{"rigid plastic"}, and \textit{"metal"}, which match the four predefined categories in the ZeroWaste dataset. These class-only queries contain no contextual information and serve as a minimal baseline for open-vocabulary generalization. All models are evaluated in a zero-shot setting on the ZeroWaste-f test set using COCO-style mAP metrics. Fig.~\ref{fig:ovod_classonly_results} depicts results, revealing uniformly low performance (mAP $\le$ 7.3), with the best predictions observed for cardboard, a large and visually distinctive class. In contrast, detection of \textit{rigid plastic}, \textit{soft plastic} and \textit{metal} remains particularly poor across all models, reflecting the challenge of recognizing deformable, transparent or reflective objects in cluttered industrial scenes. Further confusion-matrix analysis and prompt-level qualitative comparisons are provided in Appendix B.

\begin{figure}[htb]
    \centering
    \includegraphics[width=0.85\linewidth]{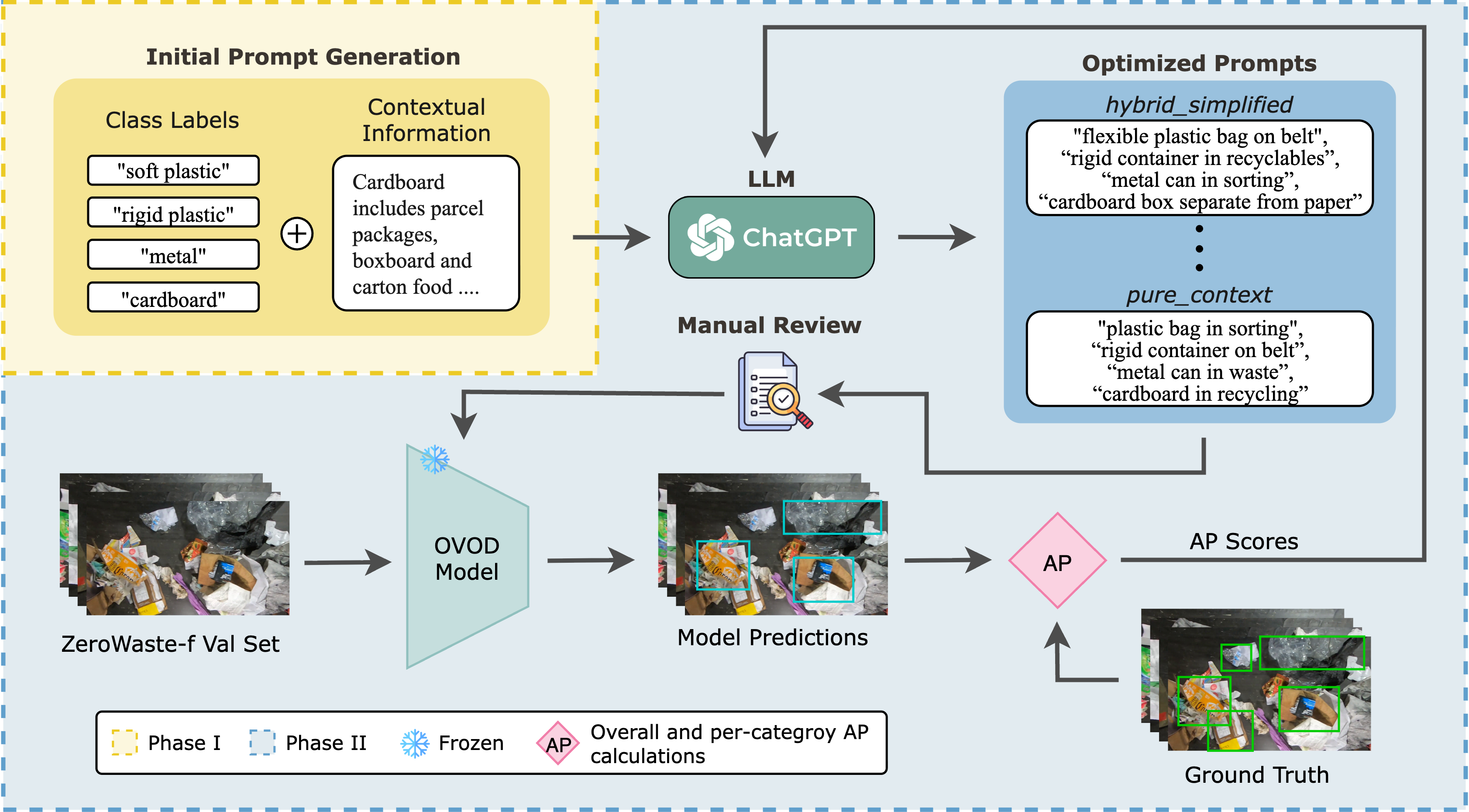}
    \caption{Iterative prompt optimization pipeline. Class-level prompts are enriched with material-specific and contextual cues using GPT-4o, then evaluated in a zero-shot setting on the ZeroWaste-f validation set using OVOD models (Grounding DINO, OWLv2). Detection performance is used to iteratively refine prompts, forming a feedback loop.}
    \label{fig:optimized_queries_pipeline} 
    \vspace{-0.6em}
\end{figure}

\begin{figure}[t]
  \centering
  \setlength{\tabcolsep}{5pt}
  \begin{tabular}{cc}
    \includegraphics[width=0.45\textwidth]{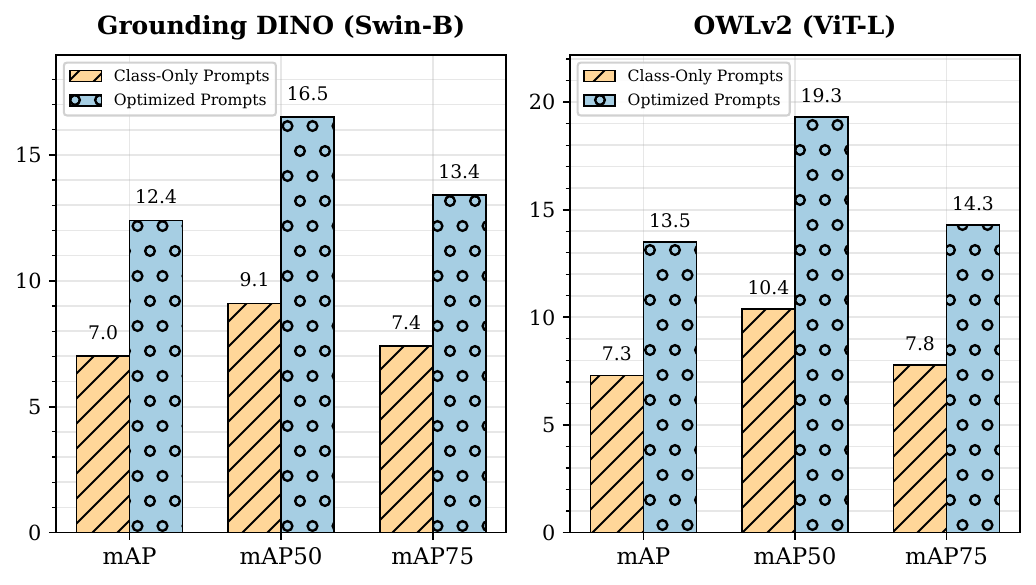} &
    \includegraphics[width=0.51\textwidth]{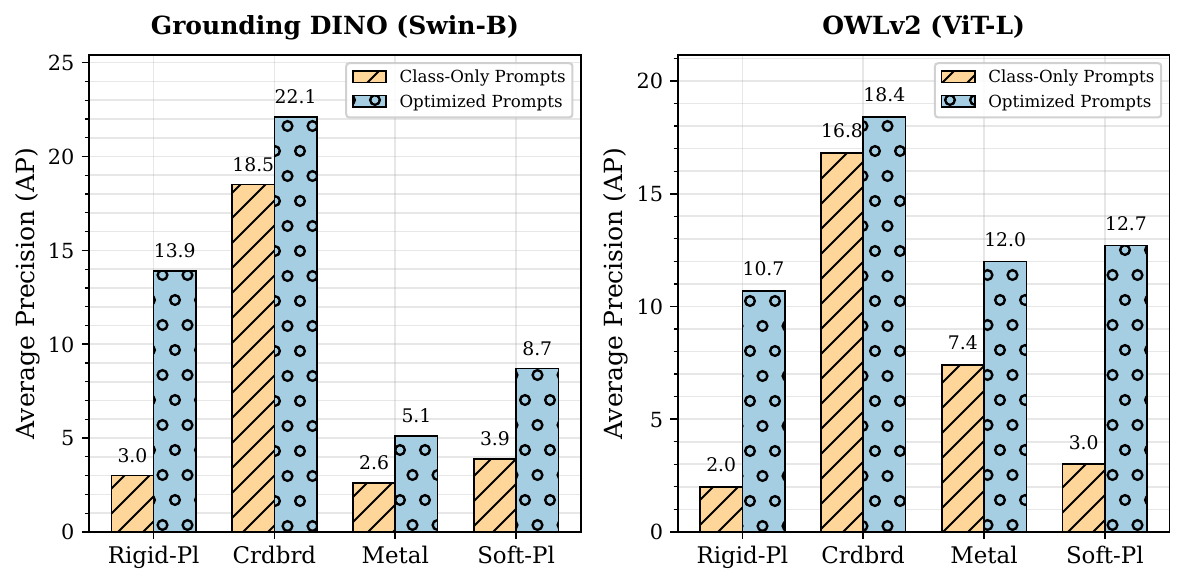} \\
    (a) Overall mAP Values & (b) Category-wise AP Values 
  \end{tabular}
  \vspace{-0.6em}
    \caption{
   Class-only vs. optimized prompts on the ZeroWaste-f test set, using Grounding DINO (Swin-B) and OWLv2 (ViT-L). (a) Overall mAP, mAP50, and mAP75 show clear gains post-optimization. (b) Per-category AP indicates substantial improvements across all categories for both models.
    }
  \label{fig:prompt_comparison_bar} \vspace{-1em}
\end{figure}

\newpage
\noindent\textbf{Prompt Optimization with LLMs.} Fig.~\ref{fig:optimized_queries_pipeline} illustrates our two-phase prompt optimization pipeline, where GPT-4o iteratively enriches class-level queries with material and contextual cues (e.g., “flexible plastic bag”). Refined prompts are evaluated on the ZeroWaste-f validation set, and the best-performing variants are used on the test set. For a detailed breakdown of prompt styles and their category-wise performance, refer to Appendix B. 
Fig.~\ref{fig:prompt_comparison_bar} shows that these optimized prompts markedly improve zero-shot performance for Grounding DINO (Swin-B) and OWLv2 (ViT-L), the top models from the class-only prompt evaluation. OWLv2 improves from 7.3 to 13.5 mAP (\textcolor{darkgreen}{+6.2}), and Grounding DINO from 7.0 to 12.4 (\textcolor{darkgreen}{+5.4}), with consistent gains across mAP50 and mAP75. All categories benefit, with the most significant improvements observed for \textit{rigid plastic} and \textit{soft plastic}. However, Grounding DINO still struggles more than OWLv2 on \textit{metal} and \textit{soft plastic}, indicating greater sensitivity to reflective and transparent materials. Qualitative examples illustrating the effects of prompt optimization on detection quality are in Appendix B.
\begin{figure}[!htp]
    \centering
    \includegraphics[width=0.95\linewidth]{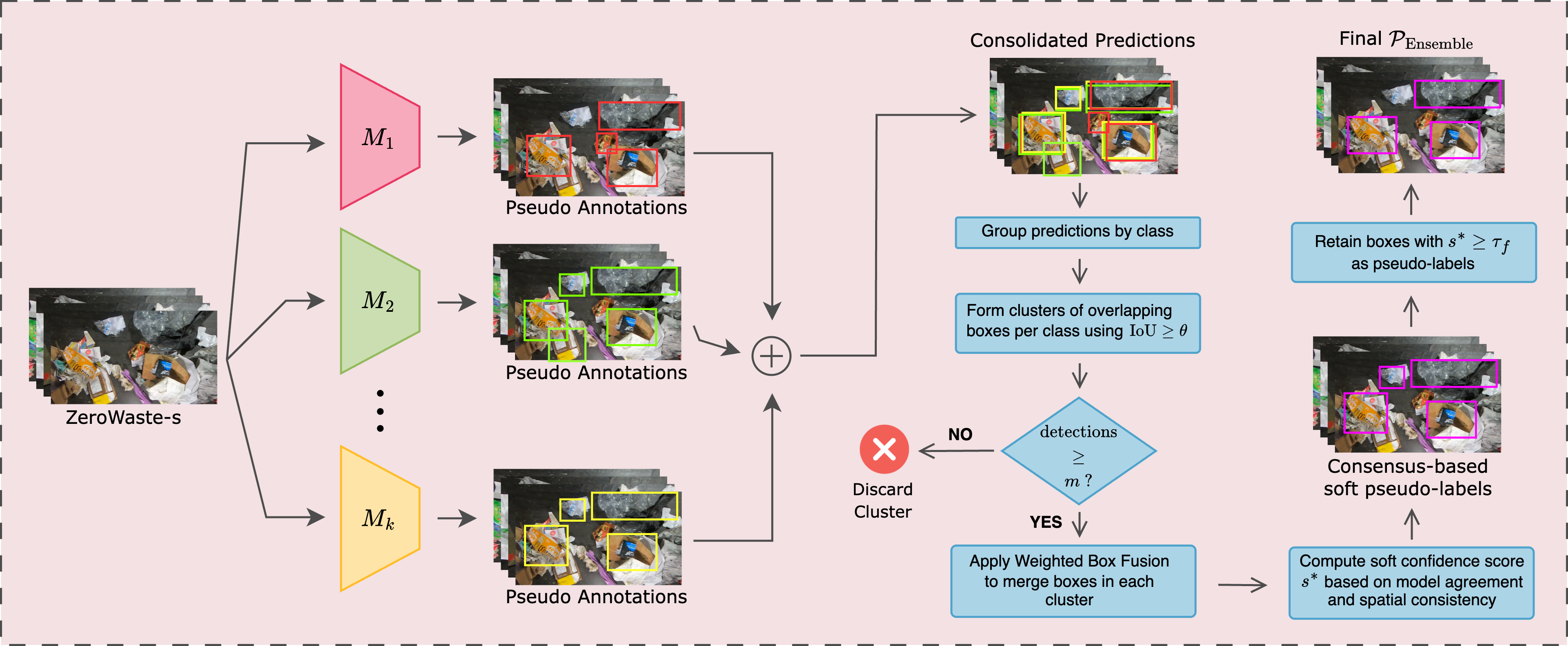}
    \caption{Soft labeling pipeline for ensemble-based pseudo-labeling. Predictions from $K$ models are filtered, clustered by class and IoU, and fused via Weighted Box Fusion. Final confidence $s^*$ is derived from base confidence and a consensus factor that accounts for spatial spread and model agreement. High-confidence pseudo-labels are retained for semi-supervised training.}
    \label{fig:pseudo_label_pipeline} \vspace{-0.6em}
\end{figure}
Despite improvements from prompt optimization, zero-shot OVOD models continue to underperform on industrial waste data, falling well below supervised baselines. Even with enriched textual prompts, these models struggle to generalize to the complex visual characteristics of industrial waste, with limitations largely driven by domain shift and insufficient exposure to waste-specific imagery during pretraining. This highlights the need for targeted task adaptation through supervised fine-tuning. For a quantitative analysis of domain shift, refer to Appendix A.

\vspace{-0.5em}
\subsection{Fine-Tuning and Updated Baselines}
To overcome the limitations of zero-shot OVOD, we establish updated baselines through fully supervised fine-tuning on ZeroWaste-f. While the original CNN-based baselines (RetinaNet \cite{lin2017focal}, Mask R-CNN~\cite{he2017mask}, and TridentNet~\cite{li2019scale}) provided a solid reference point, their performance remained low under challenging MRF conditions. Motivated by recent advances in object detection, we revisit this benchmark using modern architectures. Given the fixed label space of our task, we prioritize closed-set detectors and fine-tune six state-of-the-art models: YOLO11~\cite{jegham2025yoloevolutioncomprehensivebenchmark}, RT-DETR~\cite{Zhao_2024_CVPR}, DINO~\cite{zhang2023dino}, Co-DETR~\cite{Zong_2023_ICCV}, DETA~\cite{ouyangzhang2022nmsstrikes}, and Grounding DINO~\cite{liu2024grounding}. These models were selected for their strong performance on standard detection benchmarks and suitability for closed-set detection. YOLO11 and RT-DETR represent cutting-edge real-time architectures, while Grounding DINO, though originally designed for open-vocabulary detection, was included due to its strong zero-shot performance and adaptability to closed-set fine-tuning. For closed-set fine-tuning and inference, Grounding DINO used simple category-level prompts (\textit{"cardboard"}, \textit{"soft plastic"}, \textit{"rigid plastic"}, and \textit{"metal"}) exactly matching the dataset class names. The text encoder was frozen during fine-tuning, with only the visual backbone and detection head updated to adapt to waste detection. All other detectors in this comparison are purely visual and do not employ a text encoder, so they do not consume text prompts at inference.

We fine-tuned all models from their official pre-trained checkpoints on the ZeroWaste-f training split and evaluated on the test split using COCO metrics. Implementations followed Ultralytics~\cite{ultralytics_models} for YOLO11 and RT-DETR, MMDetection~\cite{mmdetection} for Grounding DINO, and the official repositories for DINO~\cite{zhang2023dino}, Co-DETR~\cite{Zong_2023_ICCV}, and DETA~\cite{ouyangzhang2022nmsstrikes}. Except for minibatch size, we followed each codebase’s default training recipe. Training used a single NVIDIA A100 (40 GB) GPU with batch sizes of 16 (YOLO11, RT-DETR), 4 (Grounding DINO, Swin-B), and 2 (DINO, Co-DETR, DETA, all Swin-L). As shown in Table~\ref{tab:finetuned_baselines}, fine-tuning yields substantial gains over the TridentNet baseline (24.2 mAP), with Co-DETR (Swin-L), DETA (Swin-L), and Grounding DINO (Swin-B) each achieving 51.6 mAP—more than doubling baseline performance. These results underscore the impact of task-specific fine-tuning and the superiority of transformer-based detectors over legacy CNNs and zero-shot OVOD approaches. Appendix D lists the configuration identifiers, checkpoint filenames, and pre-training datasets for all fine-tuned models.

\subsection{Semi-Supervised Learning with Ensemble-Based Pseudo-Labeling}
While fine-tuning state-of-the-art detectors on the labeled ZeroWaste-f subset ($\mathcal{D}_l$) yields strong performance, scalability remains limited by the scarcity of high-quality annotations. To address this, we leverage the unlabeled ZeroWaste-s subset ($\mathcal{D}_u$) within a semi-supervised learning (SSL) framework. Our method constructs an ensemble $\mathcal{M} = \{M_1, M_2, \dots, M_K\}$ of fine-tuned detectors and fuses their predictions to generate soft pseudo-labels. Each pseudo-label’s confidence is refined based on spatial consistency and inter-model agreement, producing reliable supervision for $\mathcal{D}_u$ without additional manual effort (Fig.~\ref{fig:pseudo_label_pipeline}). See Appendix C for an algorithmic description and qualitative comparison with human annotations.

\begin{table}[H]
  \centering
  \scriptsize
    \begin{tabular}{l|l|l|cccccc}
      \toprule
      \textbf{Model} 
        & \textbf{Backbone} 
        & \textbf{Venue}
          & \textbf{mAP} 
            & \textbf{mAP50} 
              & \textbf{mAP75} 
                & \textbf{mAPs} 
                  & \textbf{mAPm} 
                    & \textbf{mAPl} \\
      \cmidrule(lr){1-1}
      \cmidrule(lr){2-2}
      \cmidrule(lr){3-3}
      \cmidrule(lr){4-9}
        \rowcolor{mygray} 
        TridentNet & ResNet-50 & ICCV 2019 & 24.2 & 36.3 & 26.6 & 4.8 & 10.7 & 26.1 \\
        
        \rowcolor[HTML]{F6FBFF} 
        YOLO11 (L) & - & - & 34.0 {\tiny \textcolor{darkgreen}{+9.8}} & 45.3 & 36.2 & 7.3 & 16.7 & 36.6 \\
        
        \rowcolor[HTML]{F3FAFF}
        RT-DETR (L) & CSPResNet-50 & CVPR 2024 & 35.1 {\tiny \textcolor{darkgreen}{+10.9}} & 45.1 & 37.8 & 7.6 & 16.9 & 37.2 \\
        
        \rowcolor[HTML]{F1F8FF}
        DINO & ResNet-50 & ICLR 2023 & 41.0 {\tiny \textcolor{darkgreen}{+16.8}} & 51.8 & 44.4 & 9.3 & 19.9 & 44.8 \\
        
        \rowcolor[HTML]{EEF7FF}
        DINO & Swin-L & ICLR 2023 & 48.3 {\tiny \textcolor{darkgreen}{+24.1}} & 59.8 & 53.0 & 5.9 & 26.8 & 52.0 \\
        
        \rowcolor[HTML]{EBF6FF}
        DETA & ResNet-50 & - & 38.5 {\tiny \textcolor{darkgreen}{+14.3}} & 49.0 & 42.2 & 8.5 & 22.3 & 41.3 \\
        
        \rowcolor[HTML]{E9F4FF}
        DETA & Swin-L & - & 51.6 {\tiny \textcolor{darkgreen}{+27.4}} & 62.6 & 56.1 & 11.6 & 31.5 & 54.6 \\
        
        \rowcolor[HTML]{E6F3FF}
        Co-DETR & ResNet-50 & ICCV 2023 & 37.8 {\tiny \textcolor{darkgreen}{+13.6}} & 48.9 & 40.8 & 13.1 & 20.2 & 40.7 \\
        
        \rowcolor[HTML]{E3F2FF}
        Co-DETR & Swin-L & ICCV 2023 & 51.6 {\tiny \textcolor{darkgreen}{+27.4}} & 63.0 & 55.3 & 12.7 & 31.8 & 54.6 \\
        
        \rowcolor[HTML]{E1F0FF}
        Grounding DINO & Swin-T & ECCV 2024 & 45.6 {\tiny \textcolor{darkgreen}{+21.4}} & 56.8 & 50.1 & 16.6 & 27.3 & 48.9 \\
        
        \rowcolor[HTML]{DEEFFF}
        Grounding DINO & Swin-B & ECCV 2024 & 51.6 {\tiny \textcolor{darkgreen}{+27.4}} & 63.2 & 56.1 & 9.8 & 29.8 & 55.2 \\
        \bottomrule
    \end{tabular}
    \vspace{1.0em}
    \caption{Performance of fine-tuned object detectors on the ZeroWaste-f test set. We report mAP at standard thresholds and across object scales. TridentNet, the strongest baseline from the original ZeroWaste paper, is included for comparison. Relative mAP improvements over TridentNet are shown in \textcolor{darkgreen}{green}. (L) indicates large model variants.}
    \label{tab:finetuned_baselines} 
\end{table}

\noindent\textbf{Consensus-Based Pseudo-Label Generation.}
Each model $M_k\!\in\!\mathcal{M}$ predicts a set of detections on image $x\!\in\!\mathcal{D}_u$. These are first grouped by category, then clustered using pairwise IoU $\ge\theta$. For a category-specific cluster $\mathcal{C}$ that contains detections from at least $m$ distinct models, we apply weighted-box fusion (WBF)~\cite{Solovyev_2021}:

\begin{equation}
   B^{*}= \sum_{i=1}^{|\mathcal{C}|} w_i B_i, \qquad
   w_i=\frac{s_i}{\sum_{j=1}^{|\mathcal{C}|} s_j}
\end{equation}

\noindent where $B_i$ and $s_i$ denote the bounding box and confidence score of the $i$-th detection, and $|\mathcal{C}|$ is the number of detections in the cluster.

\noindent\textbf{Soft Labeling via Model Agreement.}  
We compute the soft confidence score $s^*$ for each fused box by combining spatial consistency and model agreement:

\begin{equation}\label{eq:spread}
\text{spread} = 1 - \frac{1}{|\mathcal{C}|}
                \sum_{i \in \mathcal{C}} \text{IoU}(B_i, B^*),
\end{equation}

\vspace{0.5em}
\begin{equation}\label{eq:cf}
\text{cf} = \exp(-\alpha \cdot \text{spread})
            \bigl[1 + \beta \,\bigl(|\mathcal{C}_{\text{models}}| - 2\bigr)\bigr],
\end{equation}

\begin{equation}
    s^* = s_{\text{base}} \cdot \text{cf}
\end{equation}

\noindent
Here, $|\mathcal{C}_{\text{models}}|$ is the number of contributing models, and $s_{\text{base}} = \max_{i \in \mathcal{C}} s_i$. 

For hard pseudo-labels, $s^*$ is set as $s_{\text{base}}$, without adjusting for spatial consistency or model agreement. Clusters with $s^* \ge \tau_f$ form the pseudo-label set $\mathcal{P}_{\text{Ensemble}}$. Parameters $\alpha$ and $\beta$ penalize spatial spread and reward model consensus, suppressing noisy predictions.

\noindent\textbf{Semi-Supervised Training.}
A detector $\mathcal{F}$ is trained on mixed mini-batches of $\mathcal{D}_l$ and pseudo-labeled $\mathcal{D}_u$ (ratio 1:2). Pseudo-labeled samples are weighted by $(s^{*})^{p}$ and the total training loss becomes:

\vspace{-0.5em}
\begin{equation}
  \mathcal{L}_{\text{unsup}}=\sum_{i}(s_i^{*})^{p}\left[\mathcal{L}_{\text{cls}}+\lambda\,\mathcal{L}_{\text{reg}}\right]
\end{equation}

\begin{equation}
  \mathcal{L}_{\text{total}}=\mathcal{L}_{\text{sup}}+\gamma\,\mathcal{L}_{\text{unsup}}
\end{equation}

\noindent\textbf{Implementation Details.}
The ensemble $\mathcal{M}$ comprises Grounding DINO (Swin-B), Co-DETR (Swin-L), DETA (Swin-L), and DINO (Swin-L), fine-tuned on $\mathcal{D}_l$. Predictions are filtered with $\tau_f = 0.35$, clustered with $\theta = 0.65$, and retained if $m \geq 2$ model agreement is observed. Remaining hyper-parameters are $\alpha\!=\!5.0$, $\beta\!=\!0.1$, $p\!=\!2.0$, $\lambda\!=\!2.0$, and $\gamma\!=\!1.0$. All experiments were conducted using Grounding DINO (Swin-B), which achieved the overall highest performance in the fully supervised setting (Table~\ref{tab:finetuned_baselines}).

\noindent\textbf{Results.}
Table~\ref{tab:ssl_comparative_table} summarizes the effect of different pseudo-labeling strategies. For both Swin-T and Swin-B backbones, hard pseudo-labels $\mathcal{P}_{\text{Swin-B}}$, generated by a single fine-tuned Grounding DINO (Swin-B) model, yield modest improvements over the fully supervised baseline (+2.1 and +0.9 mAP, respectively). Using ensemble-generated hard pseudo-labels further improves performance, but the best results are achieved with our proposed soft ensemble labeling, which boosts mAP to 49.3 for Swin-T and 54.3 for Swin-B, representing +3.7 and +2.7 mAP gains, respectively. To assess whether these gains disproportionately favor majority classes, Table~\ref{tab:ssl_comparative_table_perclass} reports per-class AP (AP@[50{:}95]): our soft ensemble improves all classes for both backbones, including the rare \textit{metal} class (+1.5 AP on Swin-T, +1.9 on Swin-B), with the largest gains observed on rigid and soft plastic.

\begin{table}[htb]
    \centering
    \renewcommand{\arraystretch}{1.10}
    \resizebox{\textwidth}{!}{%
    \begin{tabular}{l|c|c|l|ccc|ccc}
        \toprule
        \multirow{2}{*}{\textbf{Training Type}} & \multirow{2}{*}{\textbf{Label Type}} & \multirow{2}{*}{\textbf{Pseudo-Label Source}} & \multirow{2}{*}{\textbf{Backbone}} & \multicolumn{3}{c|}{\textbf{ZeroWaste-f mAP}} & \multicolumn{3}{c}{\textbf{Improvement over Supervised}} \\
        \cmidrule(lr){5-7} \cmidrule(lr){8-10}
        & & & & \textit{mAP} & \textit{mAP@50} & \textit{mAP@75} & $\Delta$ mAP & $\Delta$@50 & $\Delta$@75 \\
        \midrule
        \rowcolor{mygray} 
        Fully Supervised & Ground Truth & – & Swin-T & 45.6 & 56.8 & 50.1 & – & – & – \\
        \rowcolor[HTML]{F1F8FF}
        Semi-Supervised & Hard & $\mathcal{P}_{\text{Swin-B}}$ & Swin-T & 47.7 & 58.1 & 52.3 & +2.1 & +1.3 & +2.2 \\
        \rowcolor[HTML]{E6F3FF}
        Semi-Supervised & Hard & $\mathcal{P}_{\text{Ensemble}}$ {\footnotesize \textbf{(Ours)}} & Swin-T & 48.6 & 59.7 & 53.2 & +3.0 & +2.9 & +3.1 \\
        \rowcolor[HTML]{DEEFFF}
        Semi-Supervised & Soft & $\mathcal{P}_{\text{Ensemble}}$ {\footnotesize \textbf{(Ours)}} & Swin-T & \textbf{49.3} & \textbf{60.3} & \textbf{54.1} & \textbf{+3.7} & \textbf{+3.5} & \textbf{+4.0} \\
        \midrule
        \rowcolor{mygray} 
        Fully Supervised & Ground Truth & – & Swin-B & 51.6 & 63.2 & 56.0 & – & – & – \\
        \rowcolor[HTML]{F1F8FF}
        Semi-Supervised & Hard & $\mathcal{P}_{\text{Swin-B}}$ & Swin-B & 52.5 & 64.2 & 57.3 & +0.9 & +1.0 & +1.3 \\
        \rowcolor[HTML]{E6F3FF}
        Semi-Supervised & Hard & $\mathcal{P}_{\text{Ensemble}}$ {\footnotesize \textbf{(Ours)}} & Swin-B & 53.8 & 65.5 & 59.0 & +2.2 & +2.3 & +3.0 \\
        \rowcolor[HTML]{DEEFFF}
        Semi-Supervised & Soft & $\mathcal{P}_{\text{Ensemble}}$ {\footnotesize \textbf{(Ours)}} & Swin-B & \textbf{54.3} & \textbf{65.9} & \textbf{59.5} & \textbf{+2.7} & \textbf{+2.7} & \textbf{+3.5} \\
        \bottomrule
    \end{tabular}
    }
    \vspace{-0.4em}
    \caption{Comparison of supervised and semi-supervised training on the ZeroWaste-f test set using Grounding DINO with Swin-T and Swin-B backbones. Hard pseudo-labels from a single model ($\mathcal{P}_{\text{Swin-B}}$) offer moderate gains, while our ensemble-based soft labels yield the highest performance. Relative improvements over full supervision are shown in the rightmost columns.}
    \label{tab:ssl_comparative_table} 
\end{table}

\vspace{-0.5em}
\begin{table}[htb]
    \centering
    \scriptsize
    \resizebox{\textwidth}{!}{%
    \renewcommand{\arraystretch}{1.10}
    \begin{tabular}{l|c|c|l|cccc}
        \toprule
        \multirow{2}{*}{\textbf{Training Type}} & \multirow{2}{*}{\textbf{Label Type}} & \multirow{2}{*}{\textbf{Pseudo-Label Source}} & \multirow{2}{*}{\textbf{Backbone}} & \multicolumn{4}{c}{\textbf{ZeroWaste-f AP (per class)}} \\
        \cmidrule(lr){5-8}
        & & & & \textit{Rigid Plastic} & \textit{Cardboard} & \textit{Metal} & \textit{Soft Plastic} \\
        \midrule
        \rowcolor{mygray}
        Fully Supervised & Ground Truth & – & Swin-T
        & 48.9 & 50.2 & 36.6 & 47.8 \\
        \rowcolor[HTML]{E6F3FF}
        Semi-Supervised & Soft & $\mathcal{P}_{\text{Ensemble}}$ {\tiny \textbf{(Ours)}} & Swin-T
        & \textbf{53.3} \tiny \textcolor{darkgreen}{+4.4} & \textbf{52.9} \tiny \textcolor{darkgreen}{+2.7} & \textbf{38.1} \tiny \textcolor{darkgreen}{+1.5} & \textbf{52.9} \tiny \textcolor{darkgreen}{+5.1} \\
        \midrule
        \rowcolor{mygray}
        Fully Supervised & Ground Truth & – & Swin-B
        & 57.0 & 54.6 & 42.1 & 53.0 \\
        \rowcolor[HTML]{E6F3FF}
        Semi-Supervised & Soft & $\mathcal{P}_{\text{Ensemble}}$ {\tiny \textbf{(Ours)}} & Swin-B
        & \textbf{61.2} \tiny \textcolor{darkgreen}{+4.2} & \textbf{55.8} \tiny \textcolor{darkgreen}{+1.2} & \textbf{44.0} \tiny \textcolor{darkgreen}{+1.9} & \textbf{56.1} \tiny \textcolor{darkgreen}{+3.1} \\
        \bottomrule
    \end{tabular}
    }
    \vspace{-0.6em}
    \caption{Per-class AP on ZeroWaste\textendash f test set for Grounding DINO under fully supervised vs.\ semi\textendash supervised (soft, ensemble) training. Semi-supervised (soft, ensemble) improves performance across all classes, including rare ones such as \textit{metal} and \textit{soft plastic}.}
    \label{tab:ssl_comparative_table_perclass} 
\end{table}

\vspace{-1.5em}
\section{Final Pseudo-Annotations for ZeroWaste-s}
\label{sec:final_pseudo_annot}
Having demonstrated the effectiveness of our ensemble-based soft pseudo-labeling (PL) strategy, we use the best-performing model from Table~\ref{tab:ssl_comparative_table}, Grounding DINO (Swin-B), trained with soft $\mathcal{P}_{\text{Ensemble}}$ pseudo-labels, to generate a final set of high-quality annotations for the unlabeled ZeroWaste-s subset. The resulting pseudo-labels, denoted as $\mathcal{P}_{\text{Final}}$, are intended to support research in semi-supervised object detection and enhance the utility of ZeroWaste-s as a high-quality dataset for real-world industrial waste sorting applications, helping address the broader scarcity of annotated datasets in this domain. To ensure label reliability, we apply a conservative confidence threshold of 0.4, balancing precision and category diversity. The final $\mathcal{P}_{\text{Final}}$ comprises 33,075 bounding boxes across 6,065 images, with category-level statistics in Table~\ref{tab:final_pseudo_stats}.

\vspace{0.5em}
\begin{table}[htb]
\centering
\renewcommand{\arraystretch}{1.2}

\begin{minipage}[t]{0.29\textwidth}
    \centering
    \scriptsize
    \renewcommand{\arraystretch}{0.95}
    \begin{tabular}{l|c}
        \toprule
        \textbf{Category} & \textbf{Annotations} \\
        \midrule
        Cardboard & 21,352 \\
        Soft Plastic & 9,806 \\
        Rigid Plastic & 1,523 \\
        Metal & 394 \\
        \midrule
        \textbf{Total} & \textbf{33,075} \\
        \bottomrule
    \end{tabular}
    \vspace{0.5em}
    \caption{Distribution of final pseudo-annotations.} 
    \label{tab:final_pseudo_stats}
\end{minipage}
\hspace{0.04\textwidth}  
\begin{minipage}[t]{0.56\textwidth}
    \centering
    \small
    \resizebox{\textwidth}{!}{%
    \begin{tabular}{l|c|ccc}
        \toprule
        \textbf{Model} & \textbf{Training} & \textbf{mAP} & \textbf{mAP@50} & \textbf{mAP@75} \\
        \midrule
        YOLO11 (Large) & ZeroWaste-f & 34.0 & 45.3 & 36.2 \\
        YOLO11 (Large) & $\mathcal{P}_{\text{Final}}$ & \textbf{40.3} \footnotesize \textcolor{darkgreen}{+6.3} & \textbf{51.1} \footnotesize \textcolor{darkgreen}{+5.8} & \textbf{43.3} \footnotesize \textcolor{darkgreen}{+7.1} \\
        \midrule
        RT-DETR (Large) & ZeroWaste-f & 35.1 & 45.1 & 37.8 \\
        RT-DETR (Large) & $\mathcal{P}_{\text{Final}}$ & \textbf{39.4} \footnotesize \textcolor{darkgreen}{+4.3} & \textbf{52.1} \footnotesize \textcolor{darkgreen}{+7.0} & \textbf{42.6} \footnotesize \textcolor{darkgreen}{+4.8} \\
        \bottomrule
    \end{tabular}
    }
    \vspace{-0.7em}
    \caption{YOLO11 and RT-DETR trained on labeled (ZeroWaste-f) vs. pseudo-labeled ($\mathcal{P}_{\text{Final}}$) data, evaluated on ZeroWaste-f test set.}
    \label{tab:final_pseudo_eval}
\end{minipage}
\end{table}

\noindent\textbf{Indirect Evaluation via Model Transfer.} To evaluate the generalization quality of $\mathcal{P}_{\text{Final}}$, we fine-tune two high-performing detectors (YOLO11 and RT-DETR) exclusively on this pseudo-labeled dataset and evaluate their performance on the labeled ZeroWaste-f test set. This setup simulates a practical use case where manually labeled data is scarce, and model training depends entirely on pseudo-annotations. Table~\ref{tab:final_pseudo_eval} compares the performance of detectors trained on $\mathcal{P}_{\text{Final}}$ versus those trained on manually labeled ZeroWaste-f data. The results confirm that $\mathcal{P}_{\text{Final}}$ can serve as an effective alternative to manual annotation. YOLO11 trained on pseudo-labeled data achieves +6.3 mAP improvement over its fully supervised baseline, while RT-DETR sees a +4.3 mAP gain. The consistent improvements across mAP, mAP@50, and mAP@75 indicate not only better detection accuracy but also enhanced localization precision. These findings validate the generalization strength and practical utility of our pseudo-annotation pipeline in real-world applications.

\vspace{-0.75em}
\section{Conclusion}
\label{sec:conclusion}
\vspace{-0.5em}
We present a comprehensive framework for advancing AI-driven waste detection in industrial settings by evaluating zero-shot open-vocabulary object detectors, fine-tuning state-of-the-art models, and introducing a robust semi-supervised learning pipeline using ensemble-based soft pseudo-labeling. Our extensive experiments on the challenging ZeroWaste dataset reveal that while current OVOD models struggle in cluttered, deformable environments, targeted prompt optimization and fine-tuning yield substantial performance gains. Furthermore, our consensus-driven pseudo-labeling approach enables scalable learning from unlabeled data and produces high-quality annotations that rival or surpass fully supervised baselines when used for training. These contributions establish new benchmarks and outline a scalable path toward AI-assisted waste recovery in real-world material recovery facilities.

\vspace{-0.75em}
\section*{Acknowledgements}
\vspace{-0.5em}
This work was supported in part by Google unrestricted gift 2023. The authors are grateful for their generous support, which made this research possible.

\newpage
\bibliography{egbib}

\newpage
\appendix
 
\noindent\begin{huge} \textbf{Appendix} \vspace{4mm} \end{huge}

We provide extended analysis to support our main findings across four sections. \textbf{Appendix A} quantifies cross-dataset domain shift between natural and industrial imagery using Maximum Mean Discrepancy (MMD) and t-SNE visualizations. \textbf{Appendix B} offers a detailed breakdown of zero-shot Open-Vocabulary Object Detection (OVOD) performance, including confusion matrices, prompt optimization, and qualitative comparisons. \textbf{Appendix C} presents our ensemble-based pseudo-labeling algorithm in full and visualizes its alignment with human annotations. \textbf{Appendix D} lists the configuration identifiers, checkpoint filenames, and pre-training datasets used to initialize each fine-tuned detector to ensure reproducibility.

\vspace{-1.0em}
\section{Cross-Dataset Domain Shift: COCO vs. ZeroWaste}
To validate our hypothesis that zero-shot OVOD models underperform on ZeroWaste due to domain shift, we quantify the cross-dataset distributional divergence using MMD. These models are typically pre-trained on large-scale, general-purpose natural image datasets with image-text alignment, such as COCO, OpenImages, or LAION. However, their performance may degrade when applied to specialized domains like industrial waste, which differ significantly in visual characteristics. MMD provides a kernel-based statistical measure for comparing feature distributions in a reproducing kernel Hilbert space (RKHS). To assess distributional shift, we extract deep feature embeddings from a YOLO model pre-trained on COCO and compute MMD between random samples from COCO (train or val) and the ZeroWaste-f test set.

\begin{table}[H]
    \centering
    \footnotesize
    \begin{tabular}{l|c|c}
        \toprule
        \textbf{Run} & \textbf{COCO (Train) vs. ZeroWaste-f (Test)} & \textbf{COCO (Val) vs. ZeroWaste-f (Test)} \\
        \midrule
        Run 1 & 0.6125 & 0.6054 \\
        Run 2 & 0.6099 & 0.6068 \\
        Run 3 & 0.6082 & 0.6077 \\
        Run 4 & 0.5974 & 0.6046 \\
        Run 5 & 0.6111 & 0.6004 \\
        \midrule
        \textbf{Mean $\pm$ Std} & \textbf{0.6078 $\pm$ 0.0054} & \textbf{0.6050 $\pm$ 0.0026} \\
        \bottomrule
    \end{tabular}
    \vspace{0.8em}
    \caption{MMD scores across five runs comparing COCO (train and val) with the ZeroWaste-f test set. The consistently high inter-dataset MMD confirms a substantial domain shift between natural and industrial image domains.}
    \label{tab:mmd_results_coco}
\end{table}

\begin{figure}[htb]
  \centering
  \setlength{\tabcolsep}{4pt}
  \begin{tabular}{cc}
    \includegraphics[width=0.46\textwidth]{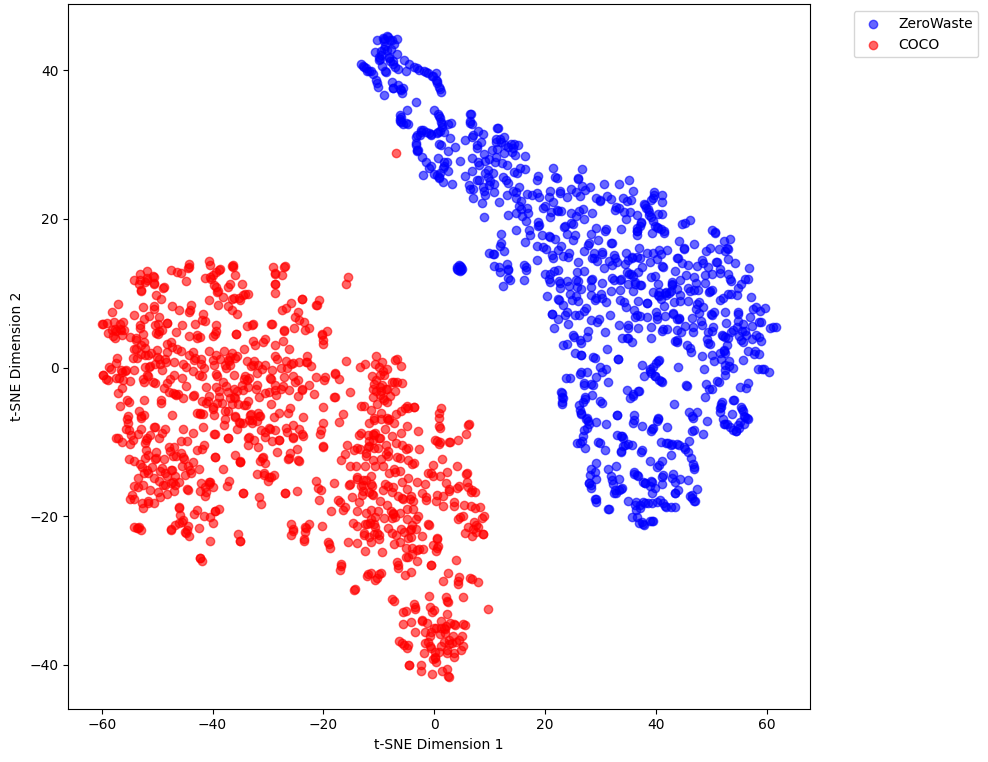} &
    \includegraphics[width=0.46\textwidth]{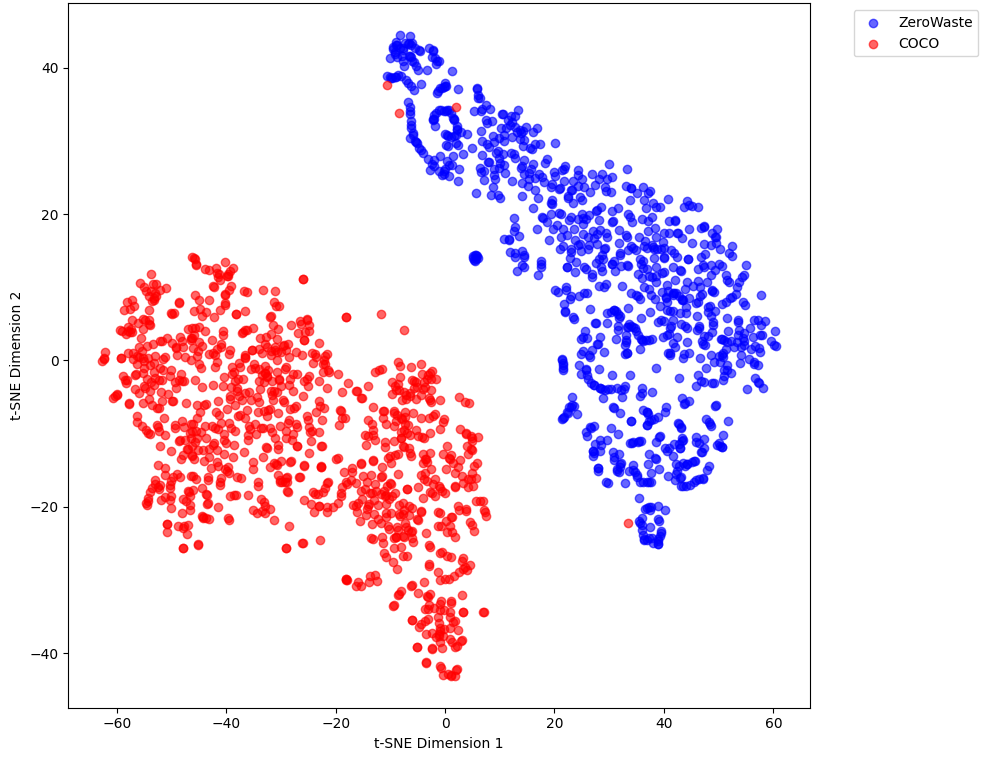} \\
    (a) COCO (Train) vs. ZeroWaste-f (Test) & (b) COCO (Val) vs. ZeroWaste-f (Test)
  \end{tabular}
  \vspace{0.4em}
    \caption{t-SNE visualizations of deep feature embeddings extracted from a YOLO model pre-trained on COCO. Red points correspond to COCO; blue points to ZeroWaste-f. The clear cluster separation in both (a) and (b) reflects a pronounced domain shift, corroborating the high MMD scores and explaining the performance drop in zero-shot OVOD settings.}
  \label{fig:tsne_domain_shift}
\end{figure}

The consistently high MMD scores ($\sim$0.61) across both COCO train and validation splits quantitatively confirm a substantial domain gap between natural and industrial imagery. Unlike COCO's diverse, object-centric scenes, ZeroWaste contains cluttered layouts, deformable objects, and domain-specific materials, factors that hinder generalization in zero-shot OVOD models pre-trained on natural image distributions. To visualize this shift, we apply t-SNE to the deep embeddings from both datasets. As shown in Figure~\ref{fig:tsne_domain_shift}, COCO and ZeroWaste embeddings form well-separated clusters with minimal overlap, further substantiating the observed cross-domain discrepancy.

\section{Extended Analysis of Zero-Shot OVOD Performance}
\begin{figure}[htb]
  \centering
  \setlength{\tabcolsep}{4pt}
  \begin{tabular}{ccc}
    \includegraphics[width=0.31\textwidth]{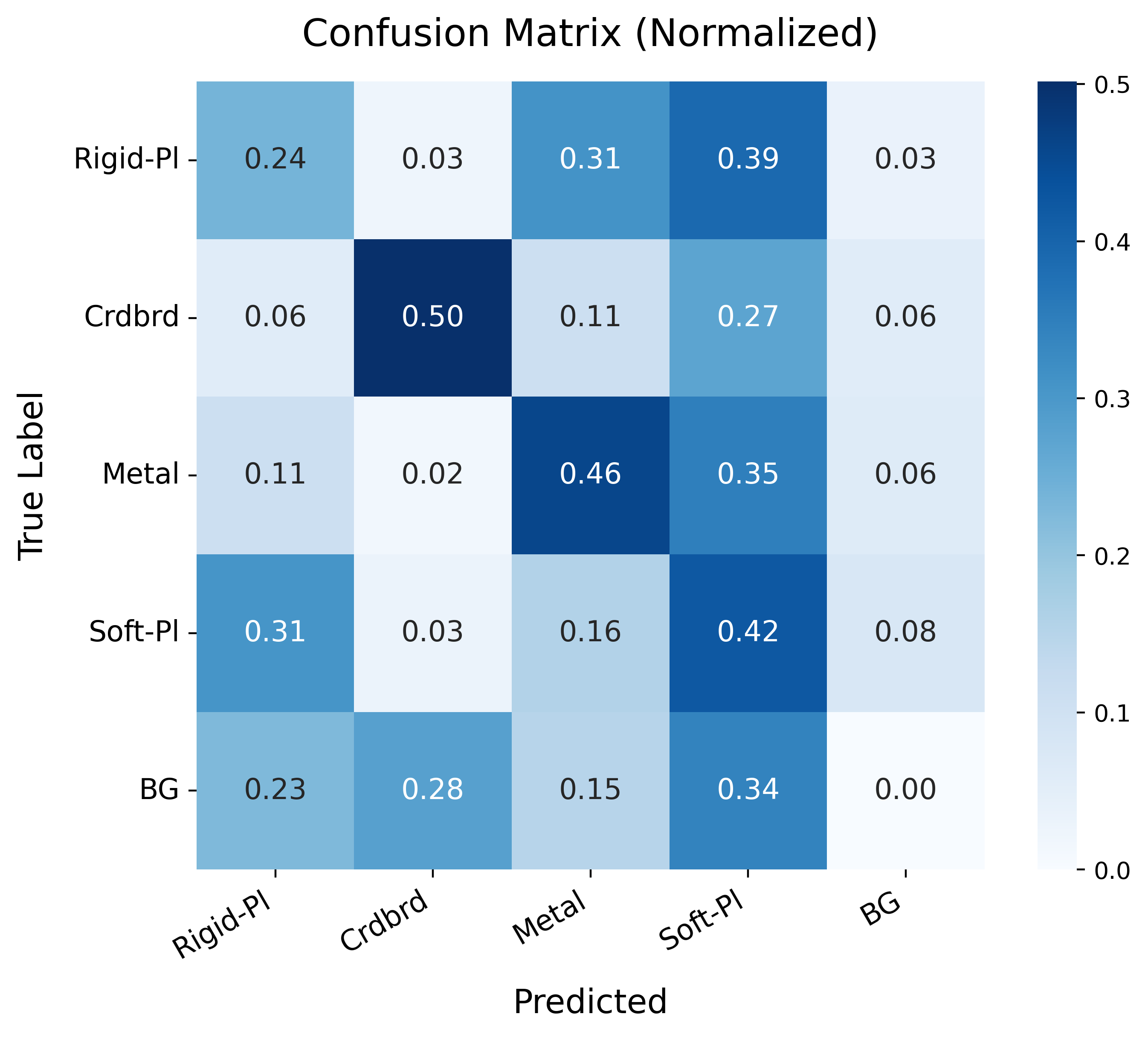} &
    \includegraphics[width=0.31\textwidth]{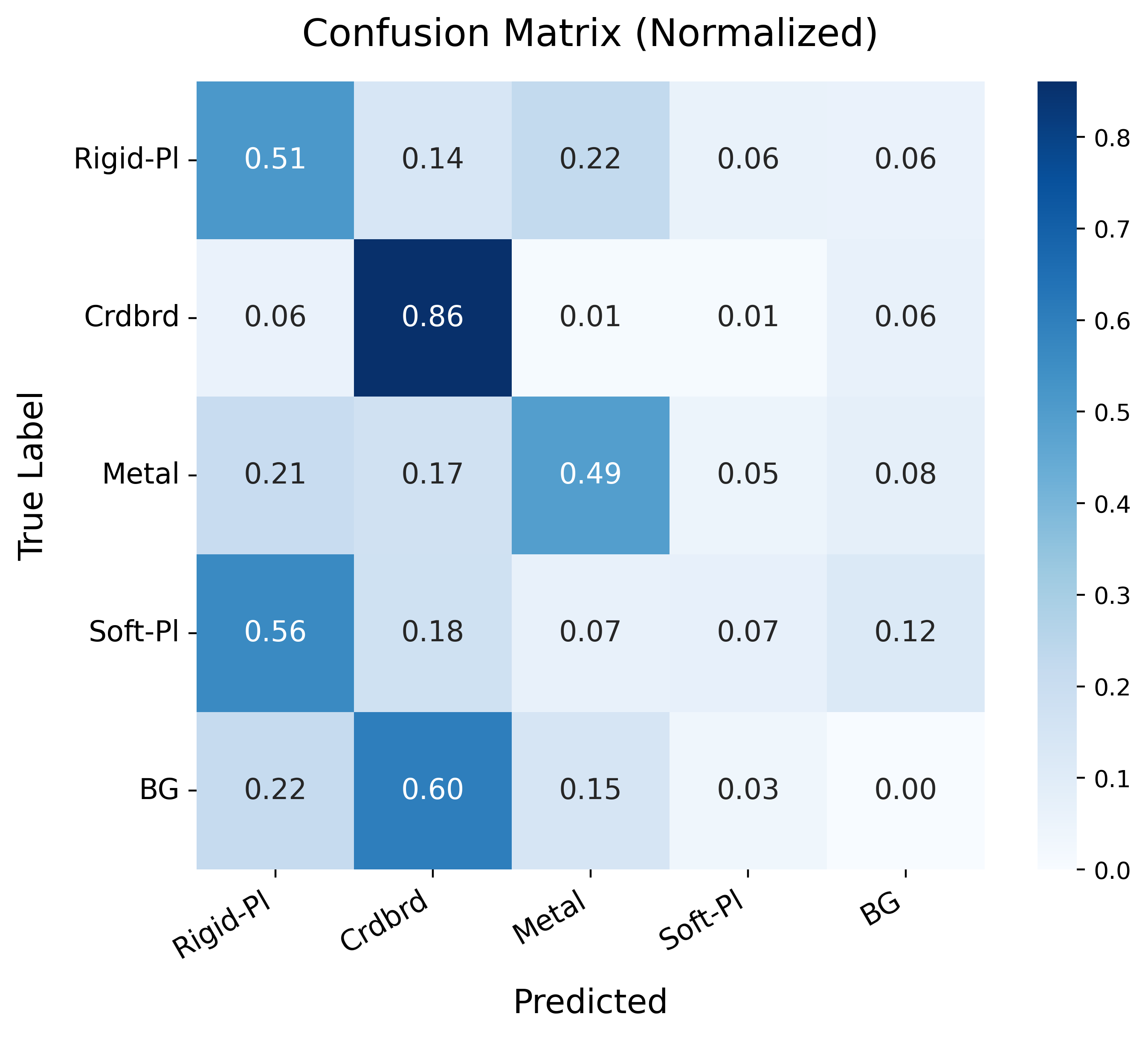} &
    \includegraphics[width=0.31\textwidth]{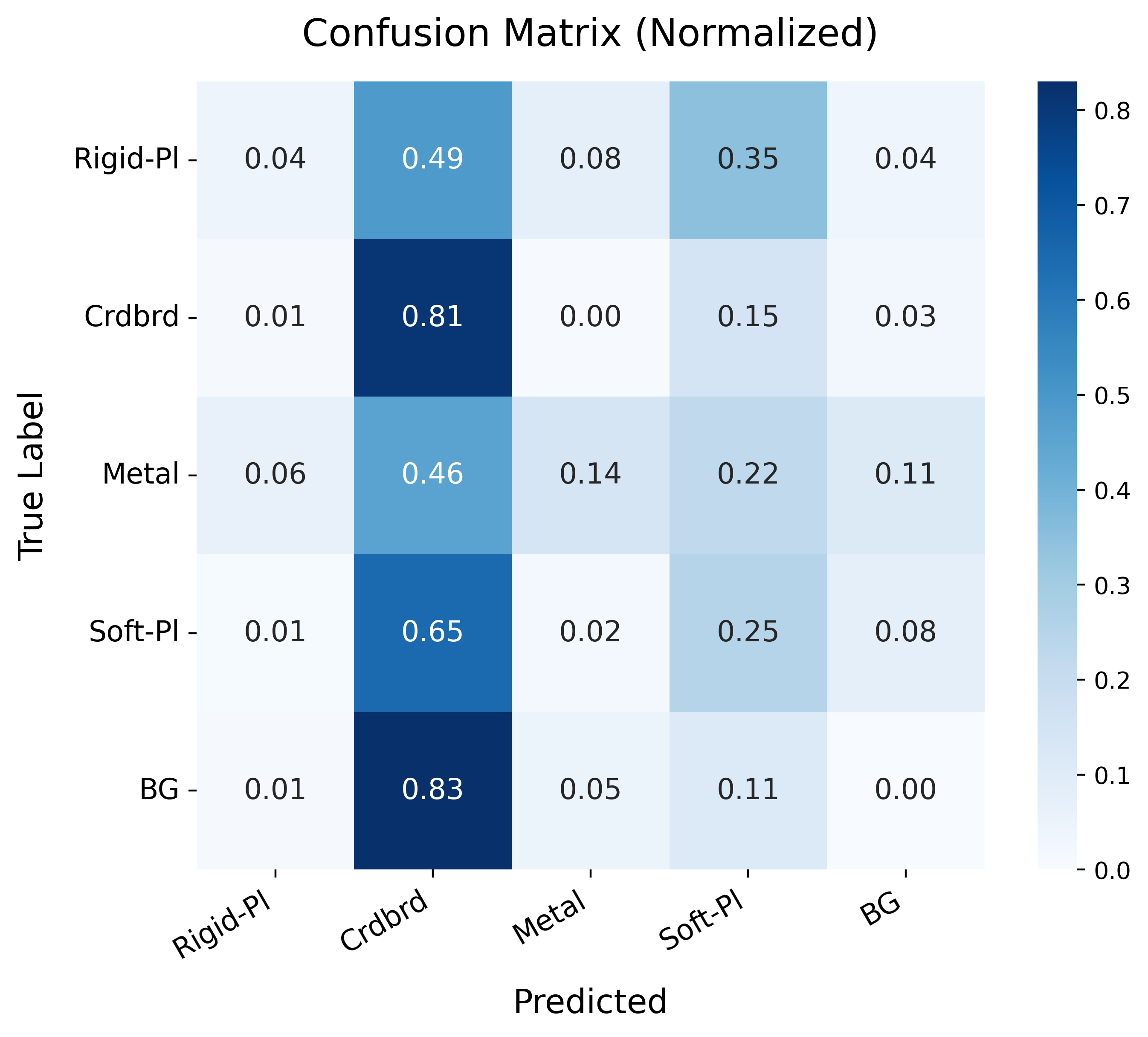} \\
    (a) Grounding DINO (Swin-B) &
    (b) OWLv2 (ViT-L) &
    (c) YOLO-World-L
  \end{tabular}
  \vspace{0.4em}
  \caption{
    Confusion matrices of zero-shot OVOD models on the ZeroWaste-f test set. \textbf{(a)} Grounding DINO shows confusion between cardboard and rigid plastic. \textbf{(b)} OWLv2 performs well on cardboard but struggles with metal. \textbf{(c)} YOLO-World-L shows greater overlap between soft plastic and rigid plastic classes.
  }
  \label{fig:confusion_matrices}
\end{figure}

\noindent\textbf{Confusion Matrix Analysis of Zero-Shot OVOD Performance.} Figure~\ref{fig:confusion_matrices} shows confusion matrices for Grounding DINO (Swin-B), OWLv2 (ViT-L), and YOLO-World-L on the ZeroWaste-f test set in a zero-shot setting. Cardboard is the most accurately detected class across models, with OWLv2 reaching 86\% accuracy. In contrast, soft plastic, rigid plastic, and cardboard are frequently confused due to their flexible, translucent, and visually similar appearances in cluttered scenes. YOLO-World shows the highest confusion, misclassifying 65\% of soft plastic as cardboard. Metal detection is the weakest across models, with over 50\% misclassification, likely due to deformation and reflective surfaces. Additionally, background regions—often containing paper and conveyor belt textures resembling cardboard—are mislabeled at high rates (60–83\%), especially by OWLv2 and YOLO-World. These trends highlight the challenges current OVOD models face in fine-grained categorization and generalization under real-world domain shift.

\noindent\textbf{Prompt Optimization Analysis.} Tables~\ref{tab:prompt_comparison_owlv2} and~\ref{tab:prompt_comparison_gdino} present a detailed breakdown of textual query styles evaluated on the ZeroWaste-f validation set for OWLv2 (ViT-L) and Grounding DINO (Swin-B). While class-only queries offer a minimal baseline (10.2 mAP), incorporating structured, context-rich descriptions significantly improves detection. The best-performing style, \textit{combined\_success}, boosts mAP to 13.3 for OWLv2 and 13.4 for Grounding DINO, demonstrating consistent gains across categories, particularly for soft and rigid plastics. Improvements stem from precise material cues (“rigid plastic container”), contextual grounding (“on conveyor belt”), and semantic redundancy (“plastic bag or wrap”). However, overly verbose or visibility-focused prompts sometimes degrade performance, especially in OWLv2, highlighting model-specific sensitivities to linguistic style. Notably, OWLv2 benefits more from spatial descriptors, while Grounding DINO responds better to concise object-centric queries. These findings emphasize the importance of prompt engineering in adapting OVOD models to complex, domain-specific settings like industrial waste sorting.

\begin{table}[H]
  \centering
  \resizebox{\textwidth}{!}{%
    \begin{tabular}{
      l
      >{\RaggedRight\arraybackslash}p{4cm}
      >{\RaggedRight\arraybackslash}p{6.5cm}
      c l | c c c c
    }
      \toprule
      \multirow{2}{*}{\textbf{Prompt Style}} &
      \multirow{2}{*}{\textbf{Description}} &
      \multirow{2}{*}{\textbf{Optimized Prompt}} &
      \multirow{2}{*}{\textbf{mAP}} &
      \multirow{2}{*}{\textbf{mAP50}} &
      \multicolumn{4}{c}{\textbf{AP per Category}} \\
      \cmidrule{6-9}
      & & & & &
      \textit{Rigid‑Pl} & \textit{Crdbrd} & \textit{Metal} & \textit{Soft‑Pl} \\
      \midrule\midrule
      
    \textbf{\textit{class-only}} 
      & (Baseline) Exact category labels
      & "soft plastic" \newline
        "rigid plastic" \newline
        "metal" \newline
        "cardboard"
      & 10.2 & 15.5 & 2.5 & 16 & 21.4 & 1.0 \\
    \midrule

    \textbf{\textit{enhanced\_properties}} 
      & Detailed material properties with location
      & "flexible translucent bag on belt" \newline
        "rigid hollow container in sorting" \newline
        "metallic can in stream" \newline
        "structured cardboard in waste"
      & 6.9 & 11.8 & 3.7 & 9.0 & 8.2 & 6.8 \\
    \midrule

    \textbf{\textit{visibility\_focused}} 
      & Combines visibility cues with context
      & "flexible plastic visible in sorting" \newline
        "rigid container visible on belt" \newline
        "metal can visible in stream" \newline
        "cardboard visible in waste"
      & 8.5 & 13.5 & 4.3 & 6.4 & 22.4 & 0.6 \\
    \midrule

    \textbf{\textit{pure\_context}} 
      & Minimal descriptors with strong context
      & "plastic bag in sorting" \newline
        "rigid container on belt" \newline
        "metal can in waste" \newline
        "cardboard in recycling"
      & 8.7 & 14.4 & 5.1 & 12.4 & 14.6 & 2.7 \\
    \midrule

    \textbf{\textit{recycling\_context}} 
      & Recycling-specific context
      & "recyclable plastic bag" \newline
        "recyclable plastic container" \newline
        "recyclable metal can" \newline
        "recyclable cardboard box"
      & 9.1 & 12.6 & 3.7 & \textbf{20.8} & 6.7 & 5.0 \\
    \midrule

    \textbf{\textit{hybrid\_simplified}} 
      & Combines successful elements with simplification
      & "flexible plastic bag on belt" \newline
        "rigid container in recyclables" \newline
        "metal can in sorting" \newline
        "cardboard box separate from \newline
        paper"
      & 9.2 & 15.8 & \textbf{7.8} & 17.9 & 10.8 & 0.4 \\
    \midrule

    \textbf{\textit{location\_enhanced}} 
      & Strong emphasis on location and sorting context
      & "plastic bag on sorting belt" \newline
        "rigid container in recycling line" \newline
        "metal can in waste stream" \newline
        "cardboard on conveyor belt"
      & 10.2 & 15.9 & 7.3 & 9.3 & 23.1 & 1.1 \\
    \midrule

    \textbf{\textit{combined\_success}} 
      & Combines elements from the most successful trials
      & "translucent plastic bag or \newline
        film in waste stream" \newline
        "sturdy rigid plastic container or \newline 
        jug in facility sorting" \newline
        "shiny metal tin or can on the \newline
        waste belt" \newline
        "brown cardboard box separated from paper"
      & \textbf{13.3} {\footnotesize \textcolor{darkgreen}{+3.1}} & \textbf{20.5} {\footnotesize \textcolor{darkgreen}{+5.0}} & 2.9 & 17.1 & \textbf{24.6} & \textbf{8.4} \\
    \bottomrule
    \end{tabular}%
  }
    \caption{Zero-shot evaluation results of OWLv2 (ViT-L) on the ZeroWaste-f validation set using different textual query styles. mAP across multiple IoU thresholds and per-category AP for rigid plastic (Rigid-Pl), cardboard (Crdbrd), metal (Metal), and soft plastic (Soft-Pl) are reported. The table demonstrates the impact of textual prompt refinement on model performance, highlighting the effectiveness of various prompt styles. The best-performing prompt style (\textit{combined\_success}) shows a relative improvement of +3.1 mAP over the baseline (class-only query). Reported improvements (\textcolor{darkgreen}{e.g., +3.1}) indicate absolute gains in overall mAP compared to the class-only prompt baseline.}
    \label{tab:prompt_comparison_owlv2}
\end{table}

\begin{center}
\vspace*{0.19\textheight}  
\begin{minipage}{0.95\textwidth}
\begin{table}[H]
  \centering
  \resizebox{\textwidth}{!}{%
    \begin{tabular}{
      l
      >{\RaggedRight\arraybackslash}p{4cm}
      >{\RaggedRight\arraybackslash}p{6.5cm}
      c l | c c c c
    }
      \toprule
      \multirow{2}{*}{\textbf{Prompt Style}} &
      \multirow{2}{*}{\textbf{Description}} &
      \multirow{2}{*}{\textbf{Optimized Prompts}} &
      \multirow{2}{*}{\textbf{mAP}} &
      \multirow{2}{*}{\textbf{mAP50}} &
      \multicolumn{4}{c}{\textbf{AP per Category}} \\
      \cmidrule{6-9}
      & & & & &
      \textit{Rigid‑Pl} & \textit{Crdbrd} & \textit{Metal} & \textit{Soft‑Pl} \\
      \midrule\midrule
      
    \textbf{\textit{class-only}} 
      & (Baseline) Exact category labels
      & \begin{minipage}[t]{6cm} \raggedright
        "soft plastic" \newline
        "rigid plastic" \newline
        "metal" \newline
        "cardboard"
        \end{minipage}
      & 10.2 & 12.8 & 5.2 & 17.9 & 15.0 & 2.8 \\
    \midrule

    \textbf{\textit{location\_enhanced}} 
      & Strong emphasis on location and sorting context
      & \begin{minipage}[t]{6cm} \raggedright
        "plastic bag among waste" \newline
        "plastic container in clutter" \newline
        "metal can on conveyor" \newline
        "cardboard box in sorting"
        \end{minipage}
      & 9.0 & 12.6 & 5.4 & 17.8 & 9.8 & 3.2 \\
    \midrule

    \textbf{\textit{pure\_context}} 
      & Minimal descriptors with strong context
      & \begin{minipage}[t]{6cm} \raggedright
        "plastic bag among waste" \newline
        "hard solid plastic container" \newline
        "metal can" \newline
        "brown cardboard box"
        \end{minipage}
      & 9.0 & 12.8 & 4.4 & 21.1 & 5.4 & 5.1 \\
    \midrule

    \textbf{\textit{recycling\_context}} 
      & Recycling-specific context
      & \begin{minipage}[t]{6cm} \raggedright
        "recyclable plastic bag" \newline
        "recyclable plastic container" \newline
        "recyclable metal can" \newline
        "recyclable cardboard box"
        \end{minipage}
      & 9.1 & 12.6 & 3.7 & 20.8 & 6.7 & 5.0 \\
    \midrule

    \textbf{\textit{hybrid\_simplified}} 
      & Combines successful elements with simplification
      & \begin{minipage}[t]{6cm} \raggedright
        "thin plastic bag" \newline
        "hard plastic container" \newline
        "metal can" \newline
        "brown cardboard box"
        \end{minipage}
      & 9.1 & 12.8 & 3.2 & 19.0 & 8.6 & 5.6 \\
    \midrule

    \textbf{\textit{enhanced\_properties}} 
      & Detailed material properties with location
      & \begin{minipage}[t]{6cm} \raggedright
        "crumpled translucent plastic bag or wrap" \newline
        "solid rigid plastic container or bottle" \newline
        "shiny reflective metal can or tin" \newline
        "thick brown cardboard box or packaging"
        \end{minipage}
      & 10.2 & 13.7 & 6.7 & 17.9 & 10.9 & 5.2 \\
    \midrule

    \textbf{\textit{visibility\_focused}} 
      & Combines visibility cues with context
      & \begin{minipage}[t]{6cm} \raggedright
        "transparent wrinkled plastic" \newline
        "hard shaped plastic" \newline
        "metallic shiny object" \newline
        "brown flat cardboard"
        \end{minipage}
      & 11.2 & 15.1 & 4.7 & 20.3 & 14.1 & 5.7 \\
    \midrule

    \textbf{\textit{combined\_success}} 
      & Combines elements from the most successful trials
      & \begin{minipage}[t]{6cm} \raggedright
        "flexible plastic bag or wrap" \newline
        "hollow rigid plastic container or bottle" \newline
        "shiny metallic can" \newline
        "stiff brown cardboard box"
        \end{minipage}
      & \textbf{13.4} {\footnotesize \textcolor{darkgreen}{+3.2}} & \textbf{18.1} {\footnotesize \textcolor{darkgreen}{+5.3}} & \textbf{8.8} & \textbf{22.1} & \textbf{16.4} & \textbf{6.3} \\
    \bottomrule
    \end{tabular}%
  }
    \caption{Zero-shot evaluation of Grounding DINO (Swin-B) with different text prompt styles under the same settings as Table \ref{tab:prompt_comparison_owlv2}. Similarly, the table highlights the impact of various prompt styles on model performance, demonstrating that the best-performing prompt style (\textit{combined\_success}) achieved a relative improvement of +3.2 mAP over the baseline (class-only prompt), as indicated in \textcolor{darkgreen}{green}. All reported gains reflect improvements in mAP relative to the class-only baseline.}
    \label{tab:prompt_comparison_gdino}
\end{table}
\end{minipage}
\end{center}

\begin{figure}[H]
  \centering
  \setlength{\tabcolsep}{4pt}
  \begin{tabular}{c}
    \includegraphics[width=0.95\textwidth]{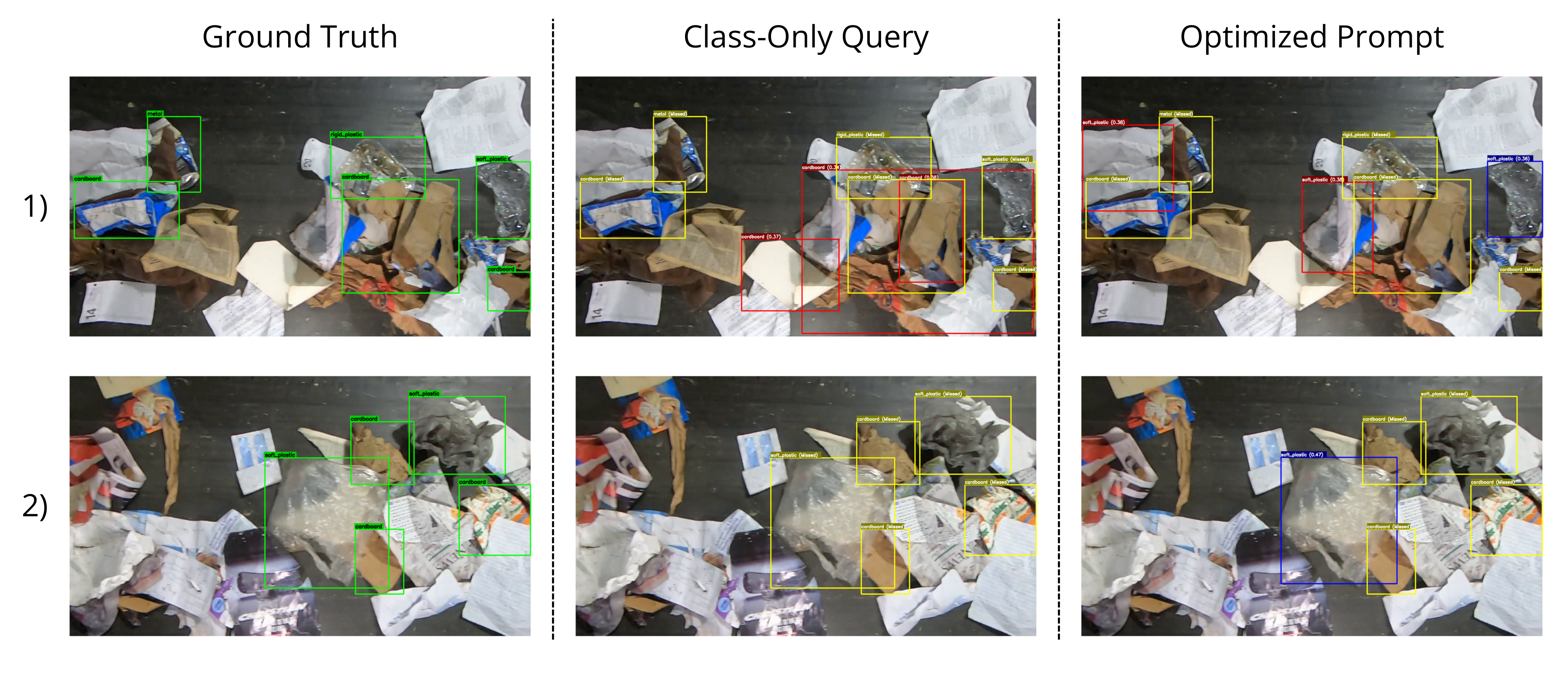} \\
    (a) OWLv2 (ViT-L) \\
    \addlinespace[0.8em]
    \includegraphics[width=0.95\textwidth]{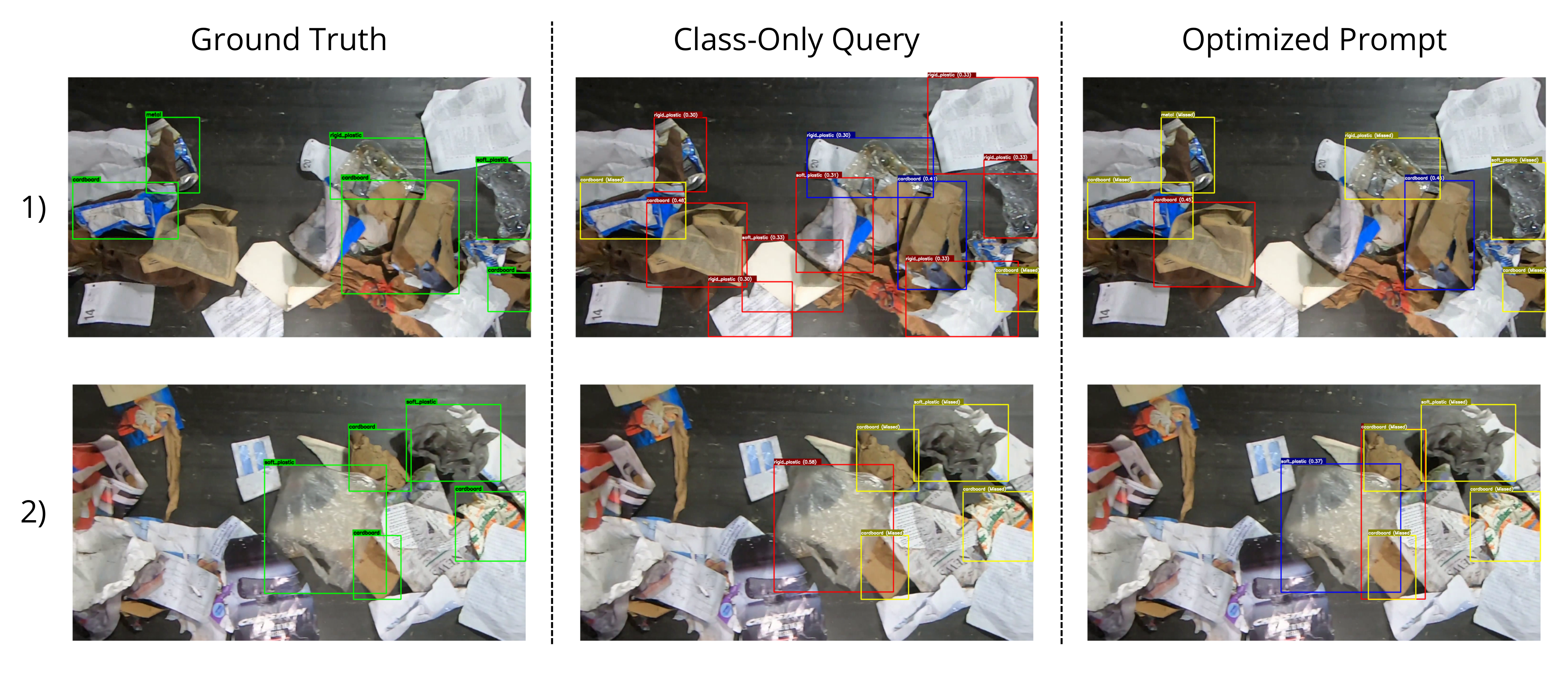} \\
    (b) Grounding DINO (Swin-B)
  \end{tabular}
  \vspace{0.4em}
  \caption{Qualitative comparison of object detection using class-only and optimized prompts on the ZeroWaste-f dataset. Ground truth annotations (green) are shown on the left, with model predictions using class-only (middle) and optimized prompts (right). Correct detections are in blue, incorrect in red, and missed detections in yellow. Optimized prompts improve localization, particularly for soft plastic and cardboard, while reducing false positives. However, both models struggle with metal detection and occluded objects. Results are shown for (a) OWLv2 (ViT-L) and (b) Grounding DINO (Swin-B).}
  \label{fig:hard_v_enhanced_annotations}
\end{figure}

\newpage
\noindent\textbf{Qualitative Results for Zero-Shot OVOD Evaluation.} Figure~\ref{fig:hard_v_enhanced_annotations} presents a visual comparison of model predictions using class-only versus optimized prompts on the ZeroWaste-f dataset. Ground truth annotations are shown on the left, while predictions from class-only and optimized prompts appear in the center and right, respectively. For both OWLv2 (ViT-L) and Grounding DINO (Swin-B), optimized prompts result in improved localization and reduced false positives, particularly for deformable classes such as soft plastic and cardboard. OWLv2 shows notable gains in recall and accuracy with enhanced prompts, while Grounding DINO exhibits reduced over-detection and cleaner bounding boxes. Nonetheless, both models consistently fail to detect metallic objects, and occluded items remain challenging. These examples illustrate the practical benefits and limitations of prompt optimization for open-vocabulary detection in cluttered, real-world waste environments.

\section{Ensemble-Based Pseudo-Labeling: Algorithm and Qualitative Analysis}
\noindent\textbf{Algorithmic Details of Ensemble-Based Pseudo-Labeling.} To complement the description in Section~3.4 of the main paper, Algorithm~\ref{alg:soft_label_pseudo_labeling} formally outlines our ensemble-based soft pseudo-labeling pipeline.

\begin{center}
\begin{minipage}{0.85\textwidth}
\begin{algorithm}[H]
    \scriptsize
    \caption{Ensemble-Based Soft Pseudo-Labeling}
    \label{alg:soft_label_pseudo_labeling}
    \begin{algorithmic}[1]
        \Require $\mathcal{M}$: Set of models $\{M_1, M_2, ..., M_k\}$, $\mathcal{D}$: Unlabeled dataset
        \Require $\tau$: Initial confidence threshold, $\theta$: IoU threshold, $m$: Minimum model agreement
        \Require $\tau_f$: Final soft confidence threshold, $\alpha$: spread decay factor, $\beta$: model agreement bonus factor
        \State \textbf{Initialize:} Pseudo-label set $\mathcal{P} \gets \emptyset$
        
        \For{each image $I \in \mathcal{D}$}
            \State \textbf{Collect Predictions:} $\mathcal{A}(I) = \bigcup_{M_i \in \mathcal{M}} M_i(I)$
            \State \textbf{Filter by Confidence:} Retain detections with $s(a) \geq \tau$
            \State \textbf{Group by Class:} $\mathcal{A}^*(I, c) = \{a \mid \text{class}(a) = c \}$
            
            \For{each category $c$}
                \State \textbf{Sort by Confidence} (highest first)
                \State \textbf{IoU Clustering:} Initialize empty clusters
                \For{each detection $a_i \in \mathcal{A}^*(I, c)$}
                    \If{$a_i$ overlaps (IoU $\geq \theta$) with an existing cluster}
                        \State Assign $a_i$ to the cluster
                    \Else
                        \State Create a new cluster with $a_i$ as reference
                    \EndIf
                \EndFor
                
                \State \textbf{Soft Weighted Box Fusion (WBF):}
                \For{each cluster $\mathcal{C}_j$}
                    \If{detections from $\geq m$ distinct models}
                        \State Compute fused bounding box:
                        \[
                        B^* = \sum_{a_i \in \mathcal{C}_j} w_i B_i, \quad w_i = \frac{s_i}{\sum s}
                        \]
                        \State Compute base confidence: $s_{\text{base}} = \max(s_i)$
                        \State Compute spread:
                        \[
                        \text{spread} = 1 - \frac{1}{K} \sum_{j=1}^{K} \text{IoU}(B_j, B^*)
                        \]
                        \State Compute consensus factor:
                        \[
                        \text{cf} = \exp(-\alpha \cdot \text{spread}) \cdot [1 + \beta \cdot (\textit{num\_models} - 2)]
                        \]
                        \State Compute soft confidence: $s^* = s_{\text{base}} \cdot \text{cf}$
                        \If{$s^* \geq \tau_f$}
                            \State Add $(B^*, s^*)$ to $\mathcal{P}$
                        \EndIf
                    \EndIf
                \EndFor
            \EndFor
        \EndFor
        
        \State \textbf{Clip Bounding Boxes} to valid image dimensions
        \State \textbf{Output:} Final soft pseudo-labels $\mathcal{P}_{\text{Ensemble}}$ in COCO format
    \end{algorithmic}
\end{algorithm}
\end{minipage}
\end{center}

\newpage
\vspace*{\fill} 

\begin{figure}[H]
    \centering
    \includegraphics[width=0.9\textwidth]{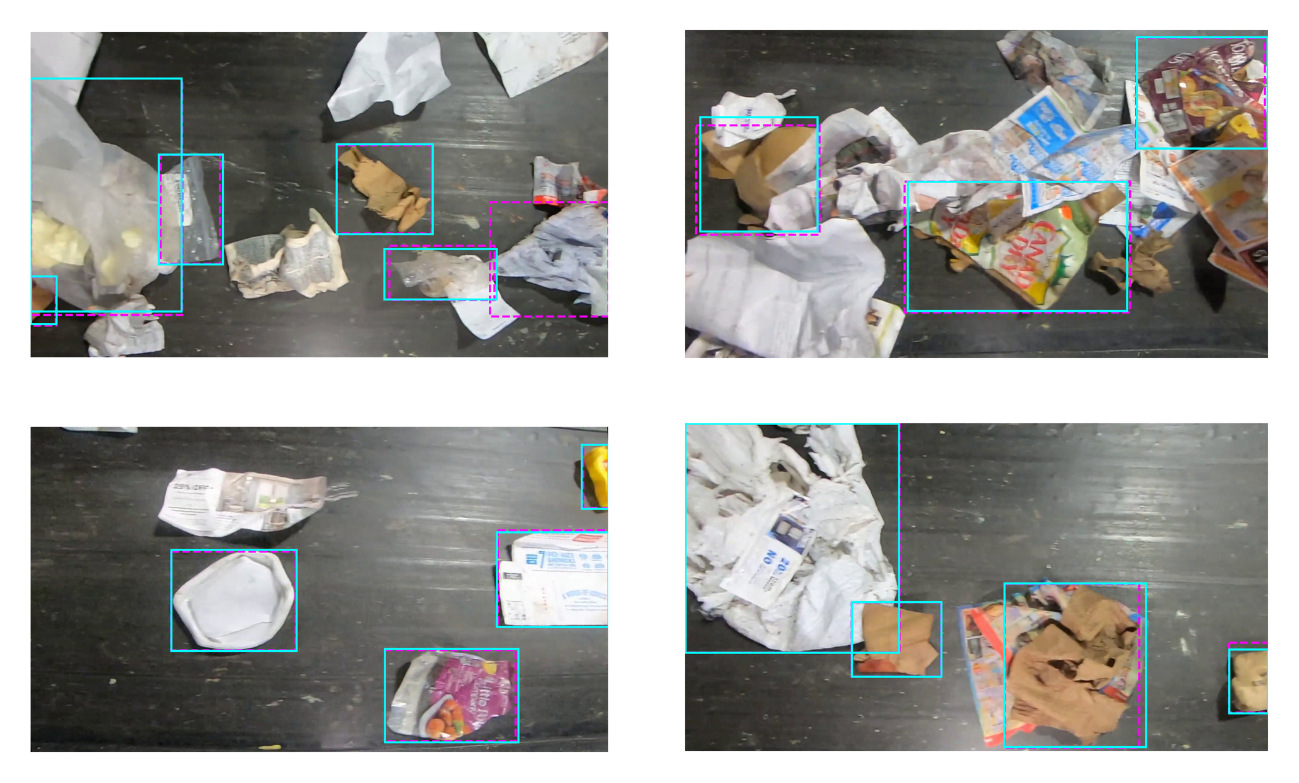}
    \caption{Qualitative comparison of fused pseudo-labels against ground-truth annotations on sample waste-sorting images. Solid cyan boxes depict our ensemble-fused detections, while dashed magenta boxes show manually-verified ground truth. Across four diverse scenes—varying object shapes, occlusions, and background clutter—the fused boxes closely align with the true annotations, demonstrating the robustness and high localization accuracy of our ensemble-based pseudo-labeling pipeline.}
    \label{fig:fused_boxes_vs_gt_boxes}
\end{figure}

\noindent\textbf{Qualitative Evaluation of Ensemble-Based Pseudo-Labels.} 
Figure~\ref{fig:fused_boxes_vs_gt_boxes} visualizes the spatial alignment between our ensemble-fused pseudo-labels and manually verified ground-truth annotations. Solid \textcolor{cyan}{cyan} boxes denote fused predictions, while dashed \textcolor{magenta}{magenta} boxes represent ground truth. The close correspondence across diverse scenes, despite occlusions, deformation, and clutter, demonstrates the high localization quality and robustness of our pseudo-labeling pipeline, reinforcing its suitability for scalable semi-supervised training.

\vspace*{\fill}
\clearpage

\newpage
\section{Fine-Tuning Initializations: Checkpoints and Pre-Training Datasets}

\noindent
All detectors were initialized from official author-released checkpoints in the cited codebases. Table~\ref{tab:finetune_init} reports, for each fine-tuned model, the backbone, codebase, config/recipe ID, checkpoint filename, and the pre-training dataset(s) documented for that checkpoint. For Grounding DINO, we used the language-aligned Swin-T and Swin-B checkpoints and froze the text encoder during fine-tuning.

\newcolumntype{Y}{>{\raggedright\arraybackslash}X}
\newcommand{\code}[1]{\begingroup\urlstyle{tt}\nolinkurl{#1}\endgroup}
\setlength{\Urlmuskip}{0mu plus 2mu}

\begin{table}[htb]
  \centering
  \scriptsize
  \setlength{\tabcolsep}{3pt}
  \renewcommand{\arraystretch}{1.40}
  \begin{tabularx}{\textwidth}{l|l|l|Y|Y|Y}
    \toprule
    \textbf{Model} & \textbf{Backbone} & \textbf{Codebase} &
    \textbf{Config ID} & \textbf{Checkpoint Filename} &
    \textbf{Pre-Training Dataset(s)} \\
    \cmidrule(lr){1-1}\cmidrule(lr){2-2}\cmidrule(lr){3-3}\cmidrule(lr){4-6}
    YOLO11 (L) & -- & Ultralytics
      & \code{model=yolo11l.pt, imgsz=640}
      & \code{yolo11l.pt}
      & COCO \\
    RT-DETR (L) & CSPResNet-50 & Ultralytics
      & \code{model=rtdetr-l.pt, imgsz=640}
      & \code{rtdetr-l.pt}
      & COCO \\
    DINO & ResNet-50 & Official repository
      & \code{DINO_4scale.py}
      & \code{checkpoint0033_4scale.pth}
      & COCO \\
    DINO & Swin-L & Official repository
      & \code{DINO_4scale_swin.py}
      & \code{checkpoint0029_4scale_swin.pth}
      & Objects365, COCO \\
    DETA & ResNet-50 & Official repository
      & \code{deta_ft}
      & \code{adet_2x_checkpoint0023.pth}
      & COCO \\
    DETA & Swin-L & Official repository
      & \code{deta_swin_ft}
      & \code{adet_swin_ft.pth}
      & Objects365, COCO \\
    Co-DETR & ResNet-50 & Official repository
      & \code{co_dino_5scale_r50_1x.py}
      & \code{co_dino_5scale_r50_1x_coco.pth}
      & COCO \\
    Co-DETR & Swin-L & Official repository
      & \code{co_dino_5scale_swin_large_16e_o365tococo.py}
      & \code{co_dino_5scale_swin_large_16e_o365tococo.pth}
      & Objects365, COCO \\
    Grounding DINO & Swin-T & MMDetection
      & \code{grounding_dino_swin-t_finetune_16xb2_1x_coco.py}
      & \code{grounding_dino_swin-t_finetune_16xb2_1x_coco_20230921_152544-5f234b20.pth}
      & Objects365, GoldG-VQA, Cap4M \\
    Grounding DINO & Swin-B & MMDetection
      & \code{grounding_dino_swin-b_finetune_16xb2_1x_coco.py}
      & \code{grounding_dino_swin-b_finetune_16xb2_1x_coco_20230921_152544-5f234b20.pth}
      & COCO, Objects365, GoldG-VQA, Cap4M, Open Images, ODinW-35, RefCOCO \\
    \bottomrule
  \end{tabularx}
  \vspace{-0.2em}
  \caption{Initialization details for fine-tuning on ZeroWaste-f. All initializations use official author-released checkpoints; the table reports the exact config/recipe and checkpoint used. Pre-training dataset(s) are as documented by the authors for each checkpoint (Grounding DINO uses language-aligned checkpoints with a frozen text encoder).}
  \label{tab:finetune_init}
\end{table}

\end{document}